\documentclass{article}
\usepackage{ijcai25}

\usepackage{times}
\usepackage{soul}
\usepackage{url}
\usepackage[hidelinks]{hyperref}
\usepackage[utf8]{inputenc}
\usepackage[small]{caption}
\usepackage{graphicx}
\usepackage{amsmath}
\usepackage{amsthm}
\usepackage{booktabs}
\usepackage{algorithm}
\usepackage[switch]{lineno}

\usepackage{latexsym}

\usepackage[utf8]{inputenc} 
\usepackage[T1]{fontenc}    
\usepackage{hyperref}       
\usepackage{url}            
\usepackage{booktabs}       
\usepackage{amsfonts}       
\usepackage{nicefrac}       
\usepackage{caption}
\usepackage{microtype}      
\usepackage{xcolor}         
\usepackage{amsmath,amssymb}
\usepackage{siunitx}
\usepackage{booktabs}
\usepackage{array}
\usepackage{multirow}

\usepackage{svg}
\usepackage{adjustbox}

\usepackage{algorithm}
\usepackage{algpseudocode}

\usepackage{setspace}

\usepackage{comment}
\usepackage{placeins}

\usepackage{cancel}

\newcommand{\citep}[1]{\cite{#1}}
\newcommand{\citet}[1]{\cite{#1}}

\usepackage[]{varioref}

\usepackage[capitalise,nameinlink]{cleveref}

\usepackage[acronym,postdot]{glossaries-extra}
\setglossarysection{section}

\newcommand{\todo}[1][]{\textcolor{red}{\bf [TODO\if\relax\detokenize{#1}\relax\else:\space#1\fi]}}

\newcommand{\chk}{\checkmark}

\setabbreviationstyle[acronym]{long-short}

\newacronym{KIPPO_slides}{KIPPO}{Koopman-Inspired\\Proximal Policy Optimization\\}

\newacronym{KIPPO}{KIPPO}{Koopman-Inspired Proximal Policy Optimization}

\newacronym{ML}{ML}{Machine Learning}
\newacronym{DL}{DL}{Deep Learning}

\newacronym{MLP}{MLP}{Multi-Layer Perceptron}
\newacronym{NN}{NN}{Neural Network}
\newacronym{DNN}{DNN}{Deep Neural Network}
\newacronym{CNN}{CNN}{Convolutional Neural Network}
\newacronym{VAE}{VAE}{Variational Autoencoder}
\newacronym{GAN}{GAN}{Generative Adversarial Network}
\newacronym{RNN}{RNN}{Recurrent Neural Network}

\newacronym{MSE}{MSE}{Mean Squared Error}
\newacronym{MAE}{MAE}{Mean Absolute Error}
\newacronym{CE}{CE}{Cross-Entropy}
\newacronym{KL}{KL}{Kullback-Leibler}

\newacronym{EWMA}{EWMA}{Exponentially Weighted Moving Average}
\newacronym{CTE}{CTE}{Cumulative Temporal Error}

\newacronym{GP}{GP}{Gaussian Process}
\newacronym{GMM}{GMM}{Gaussian Mixture Model}

\newacronym{NLP}{NLP}{Natural Language Processing}

\newacronym{RL}{RL}{Reinforcement Learning}
\newacronym{DRL}{DRL}{Deep RL}
\newacronym{MBRL}{MBRL}{Model-Based RL}
\newacronym{MFRL}{MFRL}{Model-Free RL}

\newacronym{MDP}{MDP}{Markov Decision Process}
\newacronym{POMDP}{POMDP}{Partially Observable Markov Decision Process}

\newacronym{DP}{DP}{Dynamic Programming}
\newacronym{MC}{MC}{Monte Carlo}
\newacronym{TD}{TD}{Temporal Difference}

\newacronym{GAE}{GAE}{Generalized Advantage Estimation}
\newacronym{DQN}{DQN}{Deep Q-Network}

\newacronym{VPG}{VPG}{Vanilla Policy Gradient}

\newacronym{A2C}{A2C}{Advantage Actor-Critic}
\newacronym{A3C}{A3C}{Asynchronous Advantage Actor-Critic}
\newacronym{SAC}{SAC}{Soft Actor-Critic}

\newacronym{DPG}{DPG}{Deterministic Policy Gradient}
\newacronym{DDPG}{DDPG}{Deep Deterministic Policy Gradient}
\newacronym{TD3}{TD3}{Twin Delayed DDPG}

\newacronym{TRPO}{TRPO}{Trust Region Policy Optimization}
\newacronym{PPO}{PPO}{Proximal Policy Optimization}

\newacronym{RPO}{RPO}{Robust Policy Optimization}
\newacronym{MBPO}{MBPO}{Model-Based Policy Optimization}
\newacronym{SLAC}{SLAC}{Stochastic Latent Actor-Critic}

\newacronym{CQL}{CQL}{Conservative Q-learning}

\newacronym{KT}{KT}{Koopman Theory}
\newacronym{KOT}{KOT}{Koopman Operator Theory}
\newacronym{KMD}{KMD}{Koopman Mode Decomposition}

\newacronym{DMD}{DMD}{Dynamic Mode Decomposition}
\newacronym{DMDc}{DMDc}{Dynamic Mode Decomposition with Control}
\newacronym{EDMD}{EDMD}{Extended Dynamic Mode Decomposition}
\newacronym{SINDy}{SINDy}{Sparse Identification of Nonlinear Dynamics}

\newacronym{RKHS}{RKHS}{Reproducing Kernel Hilbert Space}

\newacronym{LQR}{LQR}{Linear Quadratic Regulator}
\newacronym{PID}{PID}{Proportional-Integral-Derivative}
\newacronym{MPC}{MPC}{Model Predictive Control}

\newacronym{DKRC}{DKRC}{Deep Koopman Representation for Control}
\newacronym{KFC}{KFC}{Koopman Forward (Conservative) Q-Learning}
\newacronym{DKRL}{DKRL}{Deep Koopman Reinforcement Learning}

\newacronym{PGDK}{PGDK}{Policy Gradient with Deep Koopman Representation}
\newacronym{DKR}{DKR}{Deep Koopman Representation}

\newacronym{KARL}{KARL}{Koopman-Assisted Reinforcement Learning}
\newacronym{SKVI}{SKVI}{Soft Koopman Value Iteration}
\newacronym{SAKC}{SAKC}{Soft Actor Koopman Critic}

\newacronym{LTI}{LTI}{Linear Time-Invariant}
\newacronym{ODE}{ODE}{Ordinary Differential Equation}
\newacronym{PDE}{PDE}{Partial Differential Equation}
\newacronym{SVD}{SVD}{Singular Value Decomposition}
\newacronym{PCA}{PCA}{Principal Component Analysis}

\newacronym{NAF}{NAF}{Normalized Advantage Functions}

\newglossaryentry{reals}{
  name=\ensuremath{\mathbb{R}},
  description={set of real numbers, including all rational and irrational numbers},
}

\newglossaryentry{cumprod}{
  name=\ensuremath{\prod},
  description={cumulative product operator, computing the product of a sequence of elements: \(\prod_{i=1}^{n} a_i = a_1 \times a_2 \times \cdots \times a_n\)},
}

\newglossaryentry{expectation}{
  name=\ensuremath{\mathbb{E}},
  description={expectation operator, computing the average value of a random variable},
}

\newglossaryentry{probability}{
  name=\ensuremath{P},
  description={probability measure, assigning a value between 0 and 1 to events in a sample space, satisfying non-negativity, normalization, and countable additivity},
}

\newglossaryentry{params}{
  name=\ensuremath{\theta},
  description={trainable model parameters in machine learning or reinforcement learning, typically including weights and biases of neural networks, adjusted during learning to minimize the objective function},
}

\newglossaryentry{gradient}{
  name=\ensuremath{\nabla},
  description={gradient operator, computing the vector of partial derivatives of a scalar-valued function},
}

\newglossaryentry{nograd}{
  name=\ensuremath{\cancel{\nabla}},
  description={notation indicating that gradient computation and backpropagation are disabled for a specific operation or component},
}

\newglossaryentry{loss}{
  name=\ensuremath{\boldsymbol{\mathcal{L}}},
  description={loss function in machine learning or reinforcement learning, quantifying the discrepancy between predicted and target values, serving as an objective to be minimized during training},
}

\newglossaryentry{minibatch}{
  name=\ensuremath{\mathcal{B}},
  description={mini-batch, a subset of training data used for a single iteration of gradient descent during optimization},
}

\newglossaryentry{num-epochs}{
  name=\ensuremath{N_{\text{epochs}}},
  description={number of training epochs, i.e., the number of times the learning algorithm iterates through the entire training dataset},
}

\newglossaryentry{learning-rate}{
  name=\ensuremath{\alpha},
  description={learning rate, a hyperparameter scaling the magnitude of gradient updates applied to model parameters during optimization},
}

\newglossaryentry{loss-weight}{
  name=\ensuremath{\omega},
  description={loss weight, a hyperparameter balancing the contributions of different loss components in multi-objective optimization},
}

\newglossaryentry{state-space}{
  name=\ensuremath{\mathcal{S}},
  description={state-space, the set of all possible states that the environment can be in},
}
\newglossaryentry{action-space}{
  name=\ensuremath{\mathcal{A}},
  description={action space, the set of all possible actions that an agent can take in the environment},
}

\newglossaryentry{reward-func}{
  name=\ensuremath{R},
  description={reward function, mapping states, actions, or state-action pairs to scalar rewards},
}

\newglossaryentry{discount}{
  name=\ensuremath{\gamma},
  description={discount factor, a value between 0 and 1 determining the present value of future rewards in the return calculation},
}

\newglossaryentry{total-steps}{
  name=\ensuremath{T},
  description={total number of timesteps or interaction steps between the agent and the environment, either during a single episode (for episodic tasks) or over the entire training process (for continuing tasks)},
}

\newglossaryentry{state}{
  name=\ensuremath{x},
  description={state, an element of the state-space \(\gls{state-space}\), representing a complete description of the environment at a given timestep},
}
\newglossaryentry{action}{
  name=\ensuremath{u},
  description={an action, an element of the action space \(\gls{action-space}\), representing a decision made by the agent at a given timestep},
}
\newglossaryentry{reward}{
  name=\ensuremath{r},
  description={reward, a scalar value provided by the environment to the agent based on the current state and action, computed using the reward function \(\gls{reward-func}\)},
}
\newglossaryentry{done}{
  name=\ensuremath{d},
  description={binary flag indicating whether the current episode or rollout has terminated},
}

\newglossaryentry{trajectory}{
  name=\ensuremath{\tau},
  description={trajectory or rollout, a sequence of states, actions, and rewards generated by the agent's interaction with the environment over a finite number of timesteps},
}

\newglossaryentry{initial-state-dist}{
  name=\ensuremath{\rho_0},
  description={initial state distribution, a probability distribution over the state-space \(\gls{state-space}\) specifying the likelihood of the environment starting in each possible state},
}

\newglossaryentry{return}{
  name=\ensuremath{G},
  description={return, the discounted cumulative reward obtained by the agent from a given state onwards in a specific trajectory},
}

\newglossaryentry{expected-return}{
  name=\ensuremath{J},
  description={expected return, the average return obtained by the agent over all possible trajectories when following a policy \(\gls{policy}\)},
}

\newglossaryentry{state-val}{
  name=\ensuremath{V},
  description={state-value function, estimating the expected return starting from a given state \(\gls{state}\) and following a specific policy \(\gls{policy}\) thereafter},
}

\newglossaryentry{action-val}{
  name=\ensuremath{Q},
  description={action-value function, estimating the expected return starting from a given state \(\gls{state}\), taking a specific action \(\gls{action}\), and following a specific policy \(\gls{policy}\) thereafter},
}
\newglossaryentry{advantage}{
  name=\ensuremath{A},
  description={advantage function, quantifying the relative advantage of taking a specific action \(\gls{action}\) in a given state \(\gls{state}\) compared to the average action under the current policy \(\gls{policy}\)},
}

\newglossaryentry{policy}{
  name=\ensuremath{\pi},
  description={policy, either deterministic or stochastic, mapping states \(\gls{state}\) to actions \(\gls{action}\) or probability distributions over actions, representing the agent's decision-making strategy},
}

\newglossaryentry{dyn-trans-func}{
  name=\ensuremath{\mathbf{F}},
  description={state transition function, mapping the current state \(\gls{state}_t\) and action \(\gls{action}_t\) to the next state \(\gls{state}_{t+1}\) in a discrete-time dynamical system},
}

\newglossaryentry{sys-mat-A}{
  name=\ensuremath{\mathbf{C}},
  description={system matrix in a linear dynamical system, capturing the linear dynamics of the system states without control inputs},
}

\newglossaryentry{sys-mat-B}{
  name=\ensuremath{\mathbf{D}},
  description={control matrix in a linear dynamical system, representing the effect of control inputs on the system states},
}

\newglossaryentry{koop-op}{
  name=\ensuremath{\boldsymbol{\mathcal{K}}},
  description={Koopman operator, an infinite-dimensional linear operator describing the evolution of observables \(\gls{state-obs-func}\) of a dynamical system},
}

\newglossaryentry{koop-mat-state}{
  name=\ensuremath{\mathbf{K}},
  description={Koopman state-transition matrix, a finite-dimensional approximation of the Koopman operator \(\gls{koop-op}\) for autonomous systems, capturing the linear dynamics of the observables \(\gls{state-obs}\) without control inputs},
}

\newglossaryentry{koop-mat-action}{
  name=\ensuremath{\mathbf{B}},
  description={control matrix representing the effect of control inputs in the finite-dimensional approximation of the Koopman operator, capturing how actions \(\gls{action-obs}\) influence the evolution of observables \(\gls{state-obs}\)},
}

\newglossaryentry{state-obs-func}{
  name=\ensuremath{g},
  description={state observable function, mapping states \(\gls{state}\) from the original state-space \(\gls{state-space}\) to the lifted observable space: \(\gls{state-obs} = \gls{state-obs-func}(\gls{state})\)},
}
\newglossaryentry{action-obs-func}{
  name=\ensuremath{f},
  description={action observable function, mapping actions \(\gls{action}\) from the original action space \(\gls{action-space}\) to the lifted action observable space},
}

\newglossaryentry{state-obs}{
  name=\ensuremath{y},
  description={state observable, the result of applying the state observable function \(\gls{state-obs-func}\) to a state \(\gls{state}\)},
}
\newglossaryentry{action-obs}{
  name=\ensuremath{v},
  description={an action observable, the result of applying the action observable function \(\gls{action-obs-func}\) to an action \(\gls{action}\)},
}

\newglossaryentry{eigenval}{
  name=\ensuremath{\lambda},
  description={an eigenvalue of the Koopman operator \(\gls{koop-op}\), a scalar value \(\gls{eigenval}\) associated with an eigenfunction \(\gls{eigenfunc}\)},
}
\newglossaryentry{eigenfunc}{
  name=\ensuremath{\phi},
  description={an eigenfunction of the Koopman operator \(\gls{koop-op}\), a function \(\gls{eigenfunc}\) that, when acted upon by the Koopman operator, yields a scalar multiple of itself},
}
\newglossaryentry{koop-mode}{
  name=\ensuremath{\xi},
  description={Koopman mode, a vector-valued function that, represents the contribution of each eigenfunction to the observables and allow for the reconstruction of the original state from the spectral expansion, and together with the eigenfunctions \(\gls{eigenfunc}\) and eigenvalues \(\gls{eigenval}\), characterizes the dynamics in the lifted space},
}

\newglossaryentry{loss-ppo}{
name=\ensuremath{\gls*{loss}_{\text{PPO}}},
description={total loss function used in the \glsentryshort{PPO} algorithm},
}

\newglossaryentry{ppo-hyper-clip}{
  name=\ensuremath{\epsilon},
  description={clipping parameter in \glsentryshort{PPO}, limiting the ratio of the new policy to the old policy, ensuring conservative and stable policy updates},
}

\newglossaryentry{state-encoder}{
name=\ensuremath{\boldsymbol{\varphi}_{\gls*{state}}},
description={state encoder, an \glsentryshort{MLP} with tanh activation functions mapping states \(\gls{state}\) from the original state-space \(\gls{state-space}\) to the latent-space},
}
\newglossaryentry{state-decoder}{
  name=\ensuremath{\gls*{state-encoder}^{-1}},
  description={state decoder, an \glsentryshort{MLP} with tanh activation functions mapping latent states \(\gls{state-obs}\) back to the original state-space \(\gls{state-space}\)},
}

\newglossaryentry{action-encoder}{
name=\ensuremath{\boldsymbol{\varphi}_{\gls*{action}}},
description={action encoder, an \glsentryshort{MLP} with tanh activation functions mapping actions \(\gls{action}\) from the original action space \(\gls{action-space}\) to the latent action space},
}
\newglossaryentry{action-decoder}{
  name=\ensuremath{\gls*{action-encoder}^{-1}},
  description={action decoder, an \glsentryshort{MLP} with tanh activation functions mapping latent actions \(\gls{action-obs}\) back to the original action space \(\gls{action-space}\)},
}

\newglossaryentry{binary-mask}{
  name=\ensuremath{m},
  description={binary mask that indicates the validity of states in a trajectory \(\gls{trajectory}\), used to handle variable-length trajectories, with \(\gls{binary-mask}_t = 1\) denoting a valid state and \(\gls{binary-mask}_t = 0\) denoting an invalid or terminal state},
}

\newglossaryentry{stored-state-seqs}{
  name=\ensuremath{\mathbf{\gls*{state}}},
  description={storage tensor of shape (\gls{total-steps}, \gls{horizon}+1, *), used to store at each timestep the past \gls{horizon}+1 states from \(\gls{circ-buf-states}\); used as targets for computing prediction losses during optimization},
}

\newglossaryentry{stored-action-seqs}{
  name=\ensuremath{\mathbf{\gls*{action}}},
  description={storage tensor of shape (\gls{total-steps}, \gls{horizon}, *), used to store at each timestep the past \gls{horizon} actions from \(\gls{circ-buf-actions}\); used as inputs for computing prediction losses during optimization},
}

\newglossaryentry{stored-mask-seqs}{
  name=\ensuremath{\mathbf{\gls*{binary-mask}}},
  description={storage tensor of shape (\gls{total-steps}, \gls{horizon}+1), used to store the binary masks \(\gls{binary-mask}\) associated with the collected state sequences \(\gls{stored-state-seqs}\), indicating the validity of each state in the sequence},
}

\newglossaryentry{circ-buf-states}{
  name=\ensuremath{\overset{\curvearrowright}{\mathbf{\gls*{state}}}},
  description={circular buffer used to store the most recent states during rollouts, providing a sliding window of the agent's experience for constructing prediction targets, allowing efficient storage and retrieval of recent states without the need for frequent memory reallocation},
}

\newglossaryentry{circ-buf-actions}{
  name=\ensuremath{\overset{\curvearrowright}{\mathbf{\gls*{action}}}},
  description={circular buffer used to store the most recent actions during rollouts, providing a sliding window of the agent's decisions for constructing prediction inputs, allowing efficient storage and retrieval of recent actions without the need for frequent memory reallocation},
}

\newglossaryentry{circ-buf-dones}{
  name=\ensuremath{\overset{\curvearrowright}{\mathbf{\gls*{done}}}},
  description={circular buffer used to store the most recent termination flags \(\gls{done}\) during rollouts, used to compute the binary masks \(\gls{binary-mask}\) for handling variable-length trajectories, allowing efficient storage and retrieval of recent termination flags without the need for frequent memory reallocation},
}

\newglossaryentry{state-pred-obs}{
  name=\ensuremath{\hat{\gls*{state-obs}}},
  description={predicted state observable, the latent representation of a predicted future state \(\gls{state-pred}\), obtained by applying the learned system matrices \(\gls{koop-mat-state}\) and \(\gls{koop-mat-action}\) to the current latent state and action},
}

\newglossaryentry{state-pred}{
  name=\ensuremath{\hat{\gls*{state}}},
  description={sequence of predicted future states, obtained by decoding the predicted state observables \(\gls{state-pred-obs}\) back to the original state-space \(\gls{state-space}\) using the state decoder \(\gls{state-decoder}\)},
}

\newglossaryentry{loss-ki-rec}{
name=\ensuremath{\gls*{loss}_{\text{rec}}},
description={reconstruction loss, the \glsentryshort{MSE} between the original states \(\gls{state}\) and their reconstructions \(\tilde{\gls{state}}\) obtained by encoding and decoding the states using the state autoencoder},
}

\newglossaryentry{loss-ki-pred-ls}{
name=\ensuremath{\gls*{loss}_{\text{pred-ls}}},
description={latent-space prediction loss, the \glsentryshort{MSE} between the predicted latent states \(\gls{state-pred-obs}\) and the encoded target states \(\gls{state-encoder}(\gls{state})\) over the prediction horizon \(\gls{horizon}\)},
}

\newglossaryentry{loss-ki-pred-ss}{
name=\ensuremath{\gls*{loss}_{\text{pred-ss}}},
description={state-space prediction loss, the \glsentryshort{MSE} between the decoded predicted states \(\gls{state-decoder}(\gls{state-pred-obs})\) and the true target states \(\gls{state}\) over the prediction horizon \(\gls{horizon}\)},
}

\newglossaryentry{loss-ki-wsum}{
name=\ensuremath{\gls*{loss}_{\text{KI}}},
description={weighted sum of the reconstruction loss \(\gls{loss-ki-rec}\), latent-space prediction loss \(\gls{loss-ki-pred-ls}\), and state-space prediction loss \(\gls{loss-ki-pred-ss}\), where the weights \(\gls{loss-weight-ki-rec}\), \(\gls{loss-weight-ki-pred-ls}\), and \(\gls{loss-weight-ki-pred-ss}\) control the relative importance of each term},
}

\newglossaryentry{loss-kippo}{
name=\ensuremath{\gls*{loss}_{\text{KIPPO}}},
description={total loss for the \glsentryshort{KIPPO} framework, computed as the sum of the weighted Koopman-inspired loss \(\gls{loss-ki-wsum}\) and the standard \glsentryshort{PPO} loss \(\gls{loss-ppo}\), combining the objectives of representation learning and policy optimization},
}

\newglossaryentry{horizon}{
  name=\ensuremath{H},
  description={prediction horizon, a hyperparameter specifying the number of timesteps over which the latent dynamics \(\gls{koop-mat-state}\) and \(\gls{koop-mat-action}\) are unrolled to generate predicted states \(\gls{state-pred}\)},
}

\newglossaryentry{loss-weight-ki-rec}{
  name=\ensuremath{\gls*{loss-weight}_{1}},
  description={hyperparameter for the weight of the reconstruction loss \(\gls{loss-ki-rec}\)},
}
\newglossaryentry{loss-weight-ki-pred-ls}{
  name=\ensuremath{\gls*{loss-weight}_{2}},
  description={hyperparameter for the weight of the latent-space prediction loss \(\gls{loss-ki-pred-ls}\)},
}
\newglossaryentry{loss-weight-ki-pred-ss}{
  name=\ensuremath{\gls*{loss-weight}_{3}},
  description={hyperparameter for weight of the state-space prediction loss \(\gls{loss-ki-pred-ss}\)},
}

\newcommand{\kippoRollouts}{
  \caption{
    Rollouts Phase
  }\label{alg:kippo_rollouts}
  \begin{algorithmic}[1]\onehalfspacing
    \Require
    \Statex Environment with state-space \(\gls{state-space}\) and action space \(\gls{action-space}\)
    \Statex Policy \(\gls{policy}_{\gls{params}}\); Value function \(\gls{state-val}^{\gls{policy}_{\gls{params}}}\); State encoder \(\gls{state-encoder}\)
    \Statex Max steps (batch size) \(\gls{total-steps}\); Prediction horizon \(\gls{horizon}\)
    \State Initialize storage for trajectory data \(\{\gls{state}_t, \gls{action}_t, \gls{reward}_t, \gls{done}_t, \log\gls{policy}_t, \gls{state-val}_t\}_{t=0}^{\gls{total-steps}-1}\)
    \State Initialize temporary circular buffers
    \Statex \(\{\gls{circ-buf-states}_h\}_{h=0}^{\gls{horizon}+1} \gets 0\) \Comment{Recent history of states, initialized to zeroes}
    \Statex \(\{\gls{circ-buf-actions}_h\}_{h=0}^{\gls{horizon}} \gets 0\) \Comment{Recent history of actions, initialized to zeroes}
    \Statex \(\{\gls{circ-buf-dones}_h\}_{h=0}^{\gls{horizon}} \gets 1\) \Comment{Recent history of done flags, initialized to ones}
    \State Initialize storage for the corresponding histories at each step \(\{\gls{stored-state-seqs}_{t}, \gls{stored-action-seqs}_{t}, \gls{stored-mask-seqs}_{t}\}_{t=0}^{\gls{total-steps}-1}\)
    \State \(\gls{state}_0 \gets \text{env.reset()}\) \Comment{Initialize the environment and get initial state}
    \State Add \(\gls{state}_0\) to the circular buffer \(\gls{circ-buf-states}\)
    \For{\(t \gets 0\) to \(\gls{total-steps}\)} \Comment{Interact with the environment for \gls{total-steps} steps}
    \State \(\gls{state-obs}_t \gets \gls{state-encoder}(\gls{state}_t)\) \Comment{Encode the current state to obtain the latent representation}
    \State \(\gls{action}_t \sim \gls{policy}_{\gls{params}}(\cdot | \gls{state-obs}_t)\) \Comment{Sample an action from the policy}
    \State \(\log\gls{policy}_t \gets \log \gls{policy}_{\gls{params}}(\gls{action}_t | \gls{state-obs}_t)\) \Comment{Compute the action's log probability}
    \State \(\gls{state-val}_t \gets \gls{state-val}^{\gls{policy}_{\gls{params}}}(\gls{state-obs}_t)\) \Comment{Estimate the latent state value}
    \State \(\gls{state}_{t+1}, \gls{reward}_t, \gls{done}_{t+1} \gets \text{env.step}(\gls{action}_t)\) \Comment{Execute the action in the environment}
    \State Add \(\gls{state}_{t+1}, \gls{action}_t, \gls{done}_{t+1}\) to the circular buffers \(\gls{circ-buf-states}, \gls{circ-buf-actions}\), and \(\gls{circ-buf-dones}\), respectively
    \State \(\gls{stored-state-seqs}_t, \gls{stored-action-seqs}_t \gets \gls{circ-buf-states}, \gls{circ-buf-actions}\) \Comment{Store recent history as prediction sequences}
    \State \(\gls{stored-mask-seqs}_t \gets \gls{cumprod}_{h=0}^{\gls{horizon}-1} (1 - \gls{circ-buf-dones}_h)\) \Comment{Store cumulative product of recent inverted done flags}
    \EndFor
    \State \Return All stored data
  \end{algorithmic}
}

\newcommand{\kippoPrediction}{
  \caption{
    Future State Prediction for \gls{horizon}
    Steps in Latent Space
  }\label{alg:kippo_pred}
  \begin{algorithmic}[1]\onehalfspacing
    \Require
    \Statex State encoder \(\gls{state-encoder}\); Action encoder \(\gls{action-encoder}\); System matrices: \(\gls{koop-mat-state}, \gls{koop-mat-action}\)
    \Statex Prediction horizon \(\gls{horizon}\); Initial state \(\gls{state}_{0}\); Action sequence \(\gls{action}_{0:\gls{horizon}}\)
    \State \(y_0 \gets \gls{state-encoder}(\gls{state}_0)\) \Comment{Encode the initial state}
    \For{\(h \gets 0\) to \(\gls{horizon}\)} \Comment{Predict \gls{horizon} steps into the future}
    \State \(\gls{state-pred-obs}_{h+1} \gets \gls{koop-mat-state} y_{h} + \gls{koop-mat-action} \gls{action-encoder}(\gls{action}_{h})\) \Comment{Predict next state in latent-space}
    \EndFor
    \State \Return \(\gls{state-pred-obs}_{1:\gls{horizon}+1}\)
  \end{algorithmic}
}

\newcommand{\kippoOptimization}{
  \caption{
    Optimization Phase
  }\label{alg:kippo_optim}
  \begin{algorithmic}[1]\onehalfspacing
    \Require{}
    \Statex Policy \(\gls{policy}_{\gls{params}}\); Value function \(\gls{state-val}^{\gls{policy}_{\gls{params}}}\); Update epochs \(\gls{num-epochs}\), Prediction horizon \(\gls{horizon}\)
    \Statex State encoder \(\gls{state-encoder}\); State decoder \(\gls{state-decoder}\); Action encoder \(\gls{action-encoder}\); System matrices \(\gls{koop-mat-state}, \gls{koop-mat-action}\)
    \Statex Collected data \(\{\gls{state}_t, \gls{action}_t, \gls{reward}_t, \gls{done}_t, \log\gls{policy}_t, \gls{state-val}_t, \gls{stored-state-seqs}_t, \gls{stored-action-seqs}_t, \gls{stored-mask-seqs}_t\}_{t=0}^{\gls{total-steps}-1}\) from rollouts (\cref{alg:kippo_rollouts})
    \Statex Hyperparameters \(\gls{loss-weight-ki-pred-ls}, \gls{loss-weight-ki-pred-ss}, \gls{loss-weight-ki-rec}\) \Comment{PPO-specific hyperparameters omitted for brevity}
    \For{epoch \(\gets 1\) to \(\gls{num-epochs}\)}
    \State Randomly shuffle \(data\) and split into minibatches
    \For{each \(\gls{minibatch} \in data\)}
    \State Initialize \(\gls{loss-ki-pred-ls}, \gls{loss-ki-pred-ss} \gets 0\)
    \For{each \(\{\gls{stored-state-seqs}_t, \gls{stored-action-seqs}_t, \gls{stored-mask-seqs}_t\} \in \gls{minibatch}\)}
    \State \(\gls{state}_{0} \in \gls{stored-state-seqs}_t\) \Comment{Initial state for prediction}
    \State \(\gls{action}_{0:\gls{horizon}} \in \gls{stored-action-seqs}_t\) \Comment{Actions to take for \gls{horizon} steps}
    \State \(\gls{state-pred-obs}_{1:\gls{horizon}+1} \gets \text{LatentSpacePrediction}(\gls{state}_{0}, \gls{action}_{0:\gls{horizon}})\) \Comment{\gls{horizon}-step prediction (\cref{alg:kippo_pred})}
    \State \(\gls{state}_{1:\gls{horizon}+1} \in \gls{stored-state-seqs}_t\) \Comment{Target state sequence}
    \State \(\gls{binary-mask}_{1:\gls{horizon}+1} \in \gls{stored-mask-seqs}_t\) \Comment{Binary mask accounts for episode boundaries}
    \State \(\gls{loss-ki-pred-ls} \gets \gls{loss-ki-pred-ls} +
      \frac{1}{\gls{horizon}} \sum_{h=1}^{\gls{horizon}}
      \gls{binary-mask}_{h} (\gls{state-pred-obs}_h - \gls{state-encoder}(\gls{state}_{h}))^2\) \Comment{Latent-space pred.\ loss}
      \State \(\gls{loss-ki-pred-ss} \gets \gls{loss-ki-pred-ss} +
      \frac{1}{\gls{horizon}} \sum_{h=1}^{\gls{horizon}}
      \gls{binary-mask}_{h} (\gls{state-decoder}(\gls{state-pred}_h) - \gls{state}_{h})^2\) \Comment{State-space pred.\ loss}
    \EndFor
    \State \(\gls{loss-ki-pred-ls} \gets \frac{1}{|\gls{minibatch}|} \gls{loss-ki-pred-ls}\)
    \State \(\gls{loss-ki-pred-ss} \gets \frac{1}{|\gls{minibatch}|} \gls{loss-ki-pred-ss}\)
    \State \(\gls{loss-ki-rec} \gets \frac{1}{|\gls{minibatch}|} \sum_{\gls{minibatch}} (\gls{state-decoder}(\gls{state-encoder}(\gls{state})) - \gls{state})^2\) \Comment{Reconstruction loss}
    \State \(\gls{loss-ki-wsum} \gets \gls{loss-weight-ki-rec} \gls{loss-ki-rec} + \gls{loss-weight-ki-pred-ls} \gls{loss-ki-pred-ls} + \gls{loss-weight-ki-pred-ss}  \gls{loss-ki-pred-ss}\) \Comment{Weighted sum of loss components}
    \State \(\gls{loss-ppo} \gets \text{ComputePPOLoss}(\{\gls{state-encoder}^{\gls{nograd}}(\gls{state}), \gls{action}, ...\}_{\gls{minibatch}})\) \Comment{As per original implementation}
    \State \(\gls{loss-kippo} \gets \gls{loss-ki-wsum} + \gls{loss-ppo}\) \Comment{Total loss}
    \State Compute gradients of \(\gls{loss-kippo}\) with respect to model parameters
    \State Update model parameters with gradient descent
    \EndFor
    \EndFor
  \end{algorithmic}
}

\newcommand{\figArchitecture}{
  \centering
  \includesvg[
    width=0.6\linewidth
  ]{./images/architecture}
  \caption{
    The \glsentryshort{KIPPO} framework architecture.
    The state autoencoder (encoder \(\gls{state-encoder}\) and decoder \(\gls{state-decoder}\)) learns a compact latent representation of environment states.
    The action encoder \(\gls{action-encoder}\) maps actions to this feature space.
    Within the latent space, dynamics are governed by the linear state-transition matrix \(\gls{koop-mat-state}\) and control matrix \(\gls{koop-mat-action}\).
    The policy optimization algorithm operates on the encoded states \(\gls{state-obs}_t = \gls{state-encoder}(\gls{state}_t)\).
    This architecture enables the reformulation of nonlinear environments into a structure aligned with Koopman control theory Eq.~\ref{eq:koopman_control}.
  }\label{fig:kippo-architecture}
}

\newcommand{\figResultsEnvsMain}{
  \centering
  \includegraphics[
    width=0.6\linewidth
  ]{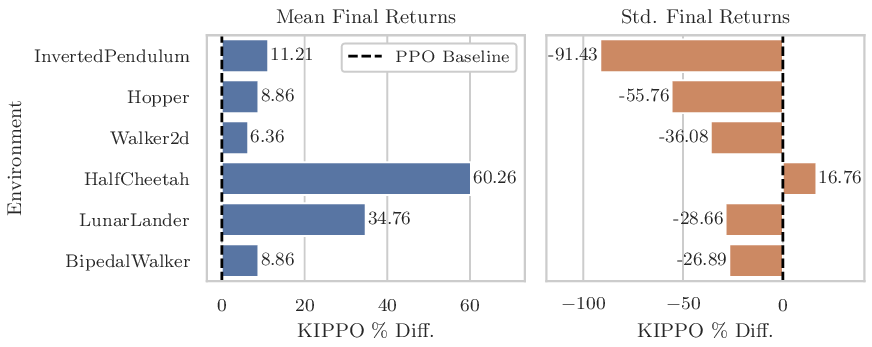}
  \caption{
    Visualization of \glsentryshort*{KIPPO}'s improvements relative to the \glsentryshort*{PPO} baseline in terms of average performance (mean, higher is better --- left) and consistency (std., lower is better
    --- right) across four trials per environment.
  }\label{fig:results_envs_main_percdiff}
}

\newcommand{\tabResultsEnvsMain}{
  \centering
  \caption{
    Per-environment overview of the main results comparing \glsentryshort{KIPPO} and the baseline \glsentryshort{PPO} and \glsentryshort{RPO} in terms of mean and std.
    of final episodic returns \gls{EWMA} across four trials.
    The bold font highlights the best performance.
  }\label{tab:results_envs_main}
  \begin{tabular}{
      l|
      rr|
      rr|
      rr
    }
    \toprule
    \textbf{Environment} & \multicolumn{2}{c}{\textbf{PPO}} & \multicolumn{2}{c}{\textbf{RPO}} & \multicolumn{2}{c}{\textbf{KIPPO}} \\
    \cmidrule(r){2-7}
                        & Mean    & Std     & Mean    & Std              & Mean              & Std \\
    \midrule
    InvertedPendulum-v4 & 897.57  & 41.61   & 892.33  & 36.46            & \bfseries 998.18  & \bfseries 3.57 \\
    Hopper-v4           & 2315.43 & 226.87  & 1970.75 & 390.81           & \bfseries 2520.53 & \bfseries 100.36 \\
    Walker2d-v4         & 3126.73 & 450.30  & 2451.58 & \bfseries 168.85 & \bfseries 3325.74 & 287.85 \\
    HalfCheetah-v4      & 1927.59 & 1030.66 & 1275.56 & \bfseries 706.52 & \bfseries 3089.20 & 1203.42 \\
    LunarLanderCont.-v2 & 208.38  & 11.59   & 150.26  & 20.14            & \bfseries 280.81  & \bfseries 8.27 \\
    BipedalWalker-v3    & 235.09  & 19.03   & 191.62  & 35.18            & \bfseries 255.91  & \bfseries 13.91 \\
    \bottomrule
  \end{tabular}

}

\newcommand{\figTrainingCurves}{
  \centering
  \includegraphics[
    width=0.8\textwidth
  ]{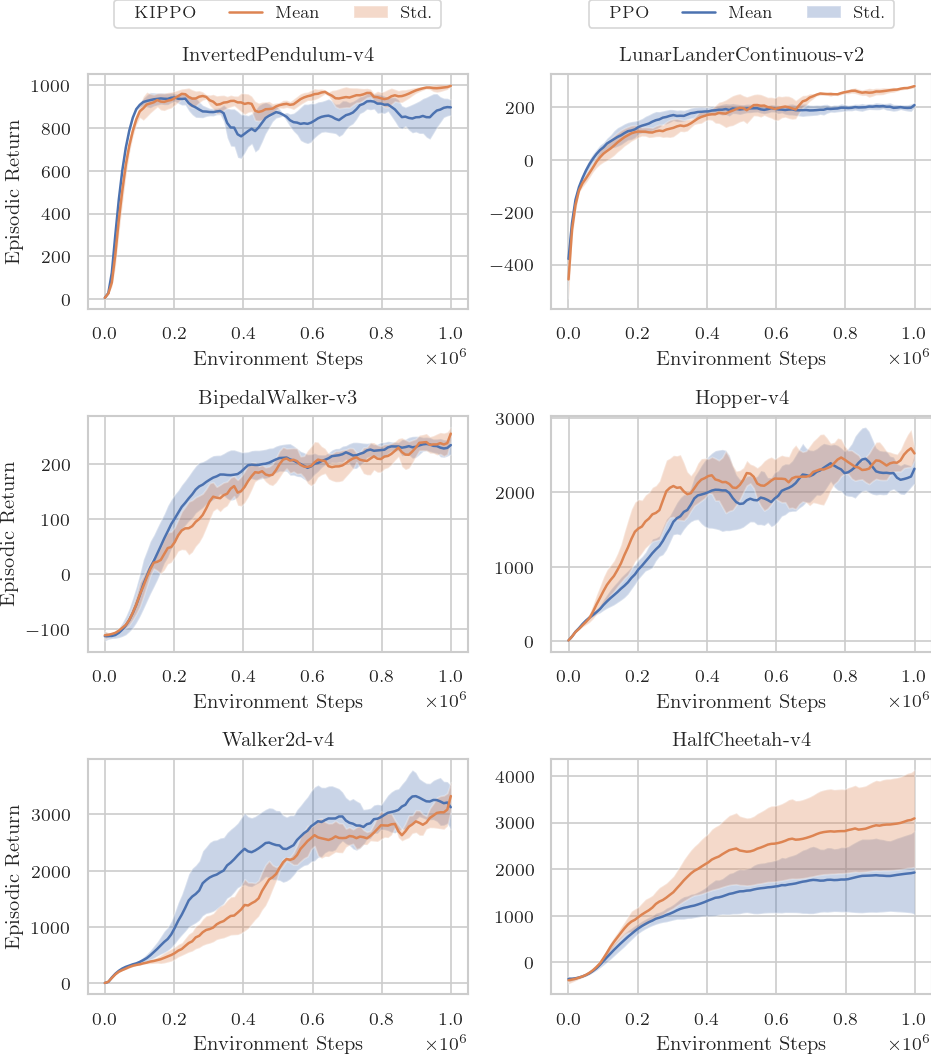}
  \caption{
    Training curves comparing \glsentryshort*{KIPPO} (orange) to the \glsentryshort*{PPO} baseline (blue) across four trials per environment.
    Solid lines represent the mean performance, while shaded regions indicate the standard deviation.
  }\label{fig:results_envs_main}
}

\newcommand{\tabEffectLosses}{
  \centering
  \caption{
    Effect of the loss function components on final episodic returns (\glsentryshort*{EWMA}) values and prediction error (\glsentryshort*{CTE}).
    The baseline is shown as the first row for each environment.
  }\label{tab:effect_losses}
  \begin{tabular}{
      l|
      ccc|
      rr|
      rr
    }
    \toprule
    \textbf{Environment} & \(\gls{loss-ki-rec}\) & \(\gls{loss-ki-pred-ls}\) & \(\gls{loss-ki-pred-ss}\) & \multicolumn{2}{c}{\textbf{\gls{EWMA}}} & \multicolumn{2}{c}{\textbf{\gls{CTE}}} \\
     &      &      &      & Mean              & Std.             & Mean            & Std. \\
    \midrule
    \multirow{0}{*}{InvertedPendulum-v4}
     & ---  & ---  & ---  & 897.57            & 41.61            & ---             & --- \\
     & \chk & ---  & ---  & 897.21            & 57.50            & 0.927           & 0.137 \\
     & ---  & \chk & ---  & 44.62             & 28.09            & 0.533           & 0.049 \\
     & ---  & ---  & \chk & 971.46            & 32.02            & \bfseries 0.001 & \bfseries 0.000 \\
     & ---  & \chk & \chk & 950.25            & 91.69            & \bfseries 0.001 & \bfseries 0.000 \\
     & \chk & \chk & ---  & 911.27            & 93.60            & 0.015           & 0.003 \\
     & \chk & ---  & \chk & 977.78            & 21.09            & \bfseries 0.001 & \bfseries 0.000 \\
     & \chk & \chk & \chk & \bfseries 998.18  & \bfseries 3.57   & \bfseries 0.001 & \bfseries 0.000 \\
    \midrule
    \multirow{0}{*}{Hopper-v4}
     & ---  & ---  & ---  & 2315.43           & 226.87           & ---             & --- \\
     & \chk & ---  & ---  & 2355.68           & 383.37           & 1.112           & 0.099 \\
     & ---  & \chk & ---  & 94.63             & 26.89            & 0.650           & 0.021 \\
     & ---  & ---  & \chk & 1583.41           & 351.23           & 0.014           & 0.002 \\
     & ---  & \chk & \chk & 2288.69           & 298.58           & 0.017           & 0.001 \\
     & \chk & \chk & ---  & 2119.36           & 139.03           & 0.066           & 0.012 \\
     & \chk & ---  & \chk & 2503.02           & 401.76           & 0.018           & \bfseries 0.003 \\
     & \chk & \chk & \chk & \bfseries 2520.53 & \bfseries 100.36 & \bfseries 0.017 & \bfseries 0.003 \\
    \midrule
    \multirow{0}{*}{Walker2d-v4}
     & ---  & ---  & ---  & 3126.73           & 450.30           & ---             & --- \\
     & \chk & ---  & ---  & 2983.97           & 374.91           & 1.325           & 0.211 \\
     & ---  & \chk & ---  & 186.95            & 62.92            & 0.339           & 0.072 \\
     & ---  & ---  & \chk & 3187.01           & 702.16           & 0.046           & 0.010 \\
     & ---  & \chk & \chk & 3196.84           & 206.52           & 0.045           & 0.005 \\
     & \chk & \chk & ---  & 2293.21           & 549.54           & 0.138           & 0.006 \\
     & \chk & ---  & \chk & 3065.78           & 585.64           & \bfseries 0.043 & 0.008 \\
     & \chk & \chk & \chk & \bfseries 3325.74 & \bfseries 287.85 & 0.054           & \bfseries 0.004 \\
    \midrule
    \multirow{0}{*}{HalfCheetah-v4}
     & ---  & ---  & ---  & 1927.59           & 1030.66          & ---             & --- \\
     & \chk & ---  & ---  & 1852.74           & 811.86           & 1.021           & 0.047 \\
     & ---  & \chk & ---  & 32.25             & 103.14           & 0.860           & 0.164 \\
     & ---  & ---  & \chk & 1608.49           & 184.02           & 0.084           & 0.005 \\
     & ---  & \chk & \chk & 2996.78           & 1100.31          & 0.107           & 0.008 \\
     & \chk & \chk & ---  & 1869.47           & \bfseries 188.03 & 0.269           & \bfseries 0.010 \\
     & \chk & ---  & \chk & 3105.47           & 1076.57          & \bfseries 0.100 & 0.011 \\
     & \chk & \chk & \chk & \bfseries 3089.20 & 1203.42          & \bfseries 0.100 & 0.011 \\
    \midrule
    \multirow{0}{*}{LunarLanderCont.-v2}
     & ---  & ---  & ---  & 208.38            & 11.59            & ---             & --- \\
     & \chk & ---  & ---  & 220.72            & 16.06            & 1.407           & 0.048 \\
     & ---  & \chk & ---  & -237.29           & 44.31            & 0.563           & 0.034 \\
     & ---  & ---  & \chk & 272.78            & 18.33            & 0.032           & 0.006 \\
     & ---  & \chk & \chk & 240.97            & 62.38            & 0.032           & 0.003 \\
     & \chk & \chk & ---  & 198.33            & 52.01            & 0.054           & 0.017 \\
     & \chk & ---  & \chk & 279.80            & \bfseries 5.14   & \bfseries 0.031 & 0.008 \\
     & \chk & \chk & \chk & \bfseries 280.81  & 8.27             & 0.036           & \bfseries  0.007 \\
    \midrule
    \multirow{0}{*}{BipedalWalker-v3}
     & ---  & ---  & ---  & 235.09            & 19.03            & ---             & --- \\
     & \chk & ---  & ---  & 233.98            & 19.58            & 1.325           & 0.115 \\
     & ---  & \chk & ---  & -126.87           & 5.70             & 0.482           & 0.097 \\
     & ---  & ---  & \chk & 268.34            & 12.46            & 0.075           & 0.010 \\
     & ---  & \chk & \chk & 241.25            & 20.89            & 0.075           & 0.007 \\
     & \chk & \chk & ---  & 214.13            & 38.04            & 0.186           & 0.026 \\
     & \chk & ---  & \chk & 235.95            & 25.78            & 0.079           & \bfseries 0.003 \\
     & \chk & \chk & \chk & \bfseries 255.91  & \bfseries 13.91  & \bfseries 0.075 & 0.007 \\
    \bottomrule
  \end{tabular}
}

\newcommand{\tabEffectLossesPartial}{
  \centering
  \caption{
    Effect of the loss function components on final episodic returns (\glsentryshort*{EWMA}) values and prediction error (\glsentryshort*{CTE}).
    The baseline is shown as the first row for each environment.
  }\label{tab:effect_losses_partial}
  \begin{tabular}{
      l|
      ccc|
      rr|
      rr
    }
    \toprule
    \textbf{Environment} & \(\gls{loss-ki-rec}\) & \(\gls{loss-ki-pred-ls}\) & \(\gls{loss-ki-pred-ss}\) & \multicolumn{2}{c}{\textbf{\gls{EWMA}}} & \multicolumn{2}{c}{\textbf{\gls{CTE}}} \\
     &      &      &      & Mean             & Std.           & Mean            & Std. \\
    \midrule
    \multirow{0}{*}{InvertedPendulum-v4}
     & ---  & ---  & ---  & 897.57           & 41.61          & ---             & --- \\
     & \chk & ---  & ---  & 897.21           & 57.50          & 0.927           & 0.137 \\
     & \chk & \chk & ---  & 911.27           & 93.60          & 0.015           & 0.003 \\
     & \chk & ---  & \chk & 977.78           & 21.09          & \bfseries 0.001 & \bfseries 0.000 \\
     & \chk & \chk & \chk & \bfseries 998.18 & \bfseries 3.57 & \bfseries 0.001 & \bfseries 0.000 \\
    \bottomrule
  \end{tabular}
}

\newcommand{\tabEffectLatentDim}{
  \centering
  \caption{
    Per-environment comparison of latent dimension effects on final episodic returns (\glsentryshort*{EWMA}) values and prediction error (\glsentryshort*{CTE}).
    The baseline is shown as the first row for each environment.
  }\label{tab:effect_latent_dim}
  \begin{tabular}{
      l|
      c|
      rr|
      rr
    }
    \toprule
    \textbf{Environment} & \textbf{Latent} & \multicolumn{2}{c}{\textbf{\gls{EWMA}}} & \multicolumn{2}{c}{\textbf{\gls{CTE}}} \\
     & \textbf{Dim.}  & Mean              & Std.             & Mean            & Std.
    \\
    \midrule
    \multirow{4}{*}{InvertedPendulum-v4}
     & ---            & 897.57            & 41.61            & ---             & --- \\
     & 16.0           & 969.94            & 48.74            & \bfseries 0.001 & \bfseries 0.000 \\
     & \bfseries 32.0 & \bfseries 998.18  & \bfseries 3.15   & \bfseries 0.001 & \bfseries 0.000 \\
     & 48.0           & 956.21            & 53.01            & \bfseries 0.001 & \bfseries 0.000 \\
    \midrule
    \multirow{4}{*}{Hopper-v4}
     & ---            & 2315.43           & 226.87           & ---             & --- \\
     & 16.0           & 1852.07           & 420.11           & 0.027           & 0.004 \\
     & 32.0           & 2246.98           & 207.23           & \bfseries 0.017 & \bfseries 0.002 \\
     & \bfseries 48.0 & \bfseries 2520.53 & \bfseries 88.78  & \bfseries 0.017 & 0.003 \\
    \midrule
    \multirow{4}{*}{Walker2d-v4}
     & ---            & 3126.73           & 450.30           & ---             & --- \\
     & 16.0           & 1728.42           & 382.02           & 0.083           & \bfseries 0.003 \\
     & 32.0           & 2565.40           & 276.60           & \bfseries 0.053 & 0.004 \\
     & \bfseries 48.0 & \bfseries 3325.74 & \bfseries 254.65 & 0.054           & \bfseries 0.003 \\
    \midrule
    \multirow{4}{*}{HalfCheetah-v4}
     & ---            & 1927.59           & 1030.66          & ---             & --- \\
     & 16.0           & 2601.19           & 968.05           & 0.147           & 0.028 \\
     & 32.0           & 1434.47           & \bfseries 575.15 & 0.117           & 0.012 \\
     & \bfseries 48.0 & \bfseries 3089.20 & 1053.22          & \bfseries 0.100 & \bfseries 0.009 \\
    \midrule
    \multirow{4}{*}{LunarLanderCont.-v2}
     & ---            & 208.38            & 11.59            & ---             & --- \\
     & 16.0           & 257.73            & 18.28            & \bfseries 0.032 & \bfseries 0.004 \\
     & 32.0           & 253.35            & 12.09            & 0.038           & 0.008 \\
     & \bfseries 48.0 & \bfseries 280.81  & \bfseries 7.32   & 0.036           & 0.006 \\
    \midrule
    \multirow{4}{*}{BipedalWalker-v3}
     & ---            & 235.09            & 19.03            & ---             & --- \\
     & 16.0           & 235.70            & \bfseries 6.48   & 0.099           & \bfseries 0.003 \\
     & 32.0           & 254.85            & 13.27            & 0.082           & 0.005 \\
     & \bfseries 48.0 & \bfseries 255.91  & 12.31            & \bfseries 0.075 & 0.006 \\
    \bottomrule
  \end{tabular}
}

\newcommand{\tabEffectHorizon}{
  \centering
  \caption{
    Per-environment comparison of prediction horizon effects on final episodic returns (\glsentryshort*{EWMA}) values and prediction error (\glsentryshort*{CTE}).
    The baseline is shown as the first row for each environment.
  }\label{tab:effect_horizon}
  \begin{tabular}{
      l|
      c|
      rr|
      rr
    }
    \toprule
    \textbf{Environment} & \textbf{Pred.}   & \multicolumn{2}{c}{\textbf{\gls{EWMA}}} & \multicolumn{2}{c}{\textbf{\gls{CTE}}} \\
     & \textbf{Horizon} & Mean              & Std.             & Mean            & Std.
    \\
    \midrule
    \multirow{5}{*}{InvertedPendulum-v4}
     & ---              & 897.57            & 41.61            & ---             & --- \\
     & 1.0              & 967.81            & 39.14            & \bfseries 0.001 & \bfseries 0.000 \\
     & \bfseries 3.0    & \bfseries 998.18  & \bfseries 3.15   & \bfseries 0.001 & \bfseries 0.000 \\
     & 5.0              & 936.10            & 67.85            & \bfseries 0.001 & \bfseries 0.000 \\
     & 10.0             & 966.83            & 39.91            & 0.003           & \bfseries 0.000 \\
    \midrule
    \multirow{5}{*}{Hopper-v4}
     & ---              & 2315.43           & 226.87           & ---             & --- \\
     & 1.0              & 2248.73           & 497.23           & \bfseries 0.014 & \bfseries 0.001 \\
     & \bfseries 3.0    & \bfseries 2520.53 & \bfseries 88.78  & 0.017           & 0.003 \\
     & 5.0              & 1824.19           & 251.75           & 0.021           & 0.002 \\
     & 10.0             & 1789.49           & 132.21           & 0.025           & 0.002 \\
    \midrule
    \multirow{5}{*}{Walker2d-v4}
     & ---              & 3126.73           & 450.30           & ---             & --- \\
     & \bfseries 1.0    & \bfseries 3325.74 & \bfseries 254.65 & \bfseries 0.054 & \bfseries 0.003 \\
     & 3.0              & 2727.15           & 434.72           & 0.066           & 0.013 \\
     & 5.0              & 2985.69           & 353.57           & 0.079           & 0.007 \\
     & 10.0             & 2965.41           & 340.65           & 0.106           & 0.007 \\
    \midrule
    \multirow{5}{*}{HalfCheetah-v4}
     & ---              & 1927.59           & 1030.66          & ---             & --- \\
     & 1.0              & 1868.40           & 697.62           & \bfseries 0.056 & 0.008 \\
     & 3.0              & 1818.68           & \bfseries 524.06 & 0.080           & 0.020 \\
     & 5.0              & 2035.49           & 1009.81          & 0.076           & \bfseries 0.007 \\
     & \bfseries 10.0   & \bfseries 3089.20 & 1053.22          & 0.100           & 0.009 \\
    \midrule
    \multirow{5}{*}{LunarLanderCont.-v2}
     & ---              & 208.38            & 11.59            & ---             & --- \\
     & 1.0              & 242.33            & 28.06            & \bfseries 0.030 & 0.007 \\
     & 3.0              & 272.58            & \bfseries 7.21   & 0.031           & \bfseries 0.005 \\
     & \bfseries 5.0    & \bfseries 280.81  & 7.32             & 0.036           & 0.006 \\
     & 10.0             & 240.80            & 48.79            & 0.045           & 0.008 \\
    \midrule
    \multirow{5}{*}{BipedalWalker-v3}
     & ---              & 235.09            & 19.03            & ---             & --- \\
     & 1.0              & 253.96            & 15.15            & \bfseries 0.059 & \bfseries 0.004 \\
     & \bfseries 3.0    & \bfseries 255.91  & 12.31            & 0.075           & 0.006 \\
     & 5.0              & 242.71            & 9.10             & 0.089           & \bfseries 0.004 \\
     & 10.0             & 233.00            & \bfseries 8.24   & 0.101           & \bfseries 0.004 \\
    \bottomrule
  \end{tabular}
}

\newcommand{\figEnvsThumbsTwoByThree}{
  \centering
  \scalebox{0.9}{
    \begin{tabular}{ccc}
      \textbf{InvertedPendulum} & \textbf{Hopper}      & \textbf{Walker} \\
      \includegraphics[width=0.3\linewidth]{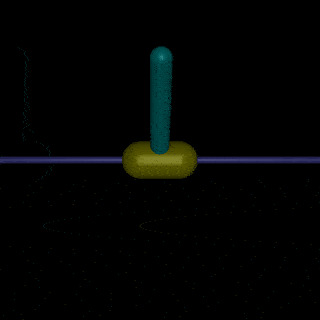} &
      \includegraphics[width=0.3\linewidth]{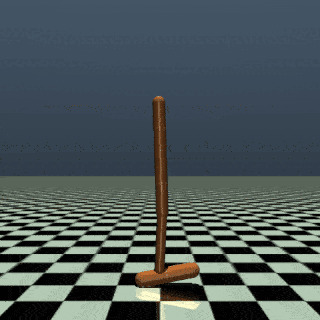}            &
      \includegraphics[width=0.3\linewidth]{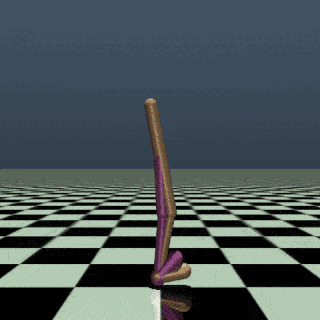} \\
      \textbf{HalfCheetah}      & \textbf{LunarLander} & \textbf{BipedalWalker} \\
      \includegraphics[width=0.3\linewidth]{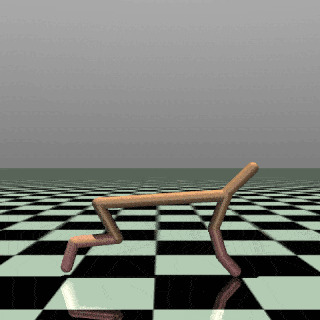} &
      \includegraphics[width=0.3\linewidth]{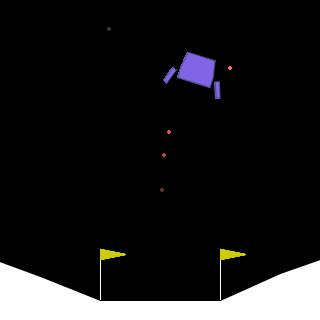} &
      \includegraphics[width=0.3\linewidth]{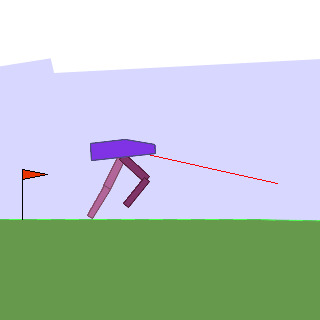}
    \end{tabular}
  }
  \caption{
    Visualizations of the six continuous control environments used for evaluation.
  }\label{fig:envs_thumbs_two_by_three}
}

\label{fig:envs_thumbs_six_by_one}

\newcommand{\tabEnvsDifficulty}{
  \centering
  \caption{
    Environments and their complexity levels based on the dimensionality of their state and action spaces.
  }\label{tab:envs_difficulty}

  \begin{tabular}{
      l
      c
      rr
    }
    \toprule
    \textbf{Environment} & \textbf{Complexity} & \textbf{|\gls{state-space}|} & \textbf{|\gls{action-space}|} \\
    \midrule
    InvertedPendulum-v4  & Low                 & 4                            & 1 \\
    LunarLanderCont.-v2  & Medium              & 8                            & 2 \\
    Hopper-v4            & Medium              & 11                           & 3 \\
    BipedalWalker-v3     & High                & 24                           & 4 \\
    Walker2d-v4          & High                & 17                           & 6 \\
    HalfCheetah-v4       & High                & 17                           & 6 \\
    \bottomrule
  \end{tabular}
}

\newcommand{\tabPPOhypers}{
  \centering
  \caption{
    Shared hyperparameters for the baseline \glsentryshort*{PPO} models and \glsentryshort*{KIPPO}.
  }\label{tab:ppo_hypers}
  \begin{tabular}{lr}
    \toprule
    \textbf{Hyperparameter}                       & \textbf{Value} \\
    \midrule
    Global Steps                                  & 1,000,000 \\
    Steps in Rollouts Phase \(\gls{total-steps}\) & 2,048 \\
    Num.
    Mini-batches                                  & 32 \\
    Num.
    Update Epochs \(\gls{num-epochs}\)            & 10 \\
    \midrule
    Discount Factor \(\gls{discount}\)            & 0.99 \\
    \glsentryshort{GAE} \(\lambda\)               & 0.95 \\
    Normalized Advantages                         & Yes \\
    Clipping Coef.
    \(\gls{ppo-hyper-clip}\)                      & 0.2 \\
    Clip Value Loss                               & Yes \\
    Weight of Value Loss                          & 0.5 \\
    Weight of Entropy Loss                        & 0.0 \\
    \midrule
    Optimizer                                     & Adam \\
    Learning Rate \(\gls{learning-rate}\)         & \(3 \times 10^{-4}\) \\
    LR Annealing                                  & Yes \\
    Max.
    Gradient Norm                                 & 0.5 \\
    \bottomrule
  \end{tabular}
}

\newcommand{\tabKIPPOhypers}{
  \centering
  \caption{
    Hyperparameter options used in the main experiments and results comparing \glsentryshort*{KIPPO} and the \glsentryshort*{PPO} baseline.
    The latent dimension and prediction horizon are varied per-environment for analysis, while the others are fixed.
  }\label{tab:kippo_hypers}
  \begin{tabular}{lr}
    \toprule
    \textbf{Hyperparameter}                                                & \textbf{Value} \\
    \midrule
    Latent Dimension                                                       & Varied (16, 32, 48) \\
    Prediction Horizon \(\gls{horizon}\)                                   & Varied (1, 3, 5, 10) \\
    \midrule
    Number of Layers                                                       & 2 \\
    Neurons per Layer                                                      & 128 \\
    \midrule
    Weight of \(\gls{loss-ki-rec}\) (\(\gls{loss-weight-ki-rec}\))         & 0.75 \\
    Weight of \(\gls{loss-ki-pred-ls}\) (\(\gls{loss-weight-ki-pred-ls}\)) & 0.1 \\
    Weight of \(\gls{loss-ki-pred-ss}\) (\(\gls{loss-weight-ki-pred-ss}\)) & 0.5 \\
    \bottomrule
  \end{tabular}
}

\newcommand{\tabHypersRanges}{
  \centering
  \caption{
    Hyperparameter values and ranges explored in the analysis.
  }\label{tab:hypers_ranges}
  \begin{tabular}{lr}
    \toprule
    \textbf{Hyperparameter}                                                & \textbf{Values/Ranges} \\
    \midrule
    Latent Dimension                                                       & 16, 32, 48, 64 \\
    Prediction Horizon \(\gls{horizon}\)                                   & 1, 3, 5, 10 \\
    \midrule
    Number of Layers                                                       & 1, 2, 3 \\
    Neurons per Layer                                                      & 64, 128, 192, 256 \\
    \midrule
    Weight of \(\gls{loss-ki-rec}\) (\(\gls{loss-weight-ki-rec}\))         & 0.00, 0.05, \ldots, 1.00 \\
    Weight of \(\gls{loss-ki-pred-ls}\) (\(\gls{loss-weight-ki-pred-ls}\)) & 0.00, 0.05, \ldots, 1.00 \\
    Weight of \(\gls{loss-ki-pred-ss}\) (\(\gls{loss-weight-ki-pred-ss}\)) & 0.00, 0.05, \ldots, 1.00 \\
    \bottomrule
  \end{tabular}
}

\newcommand{\figHypersImportance}{
  \centering
  \includegraphics[
    width=1.0\linewidth
  ]{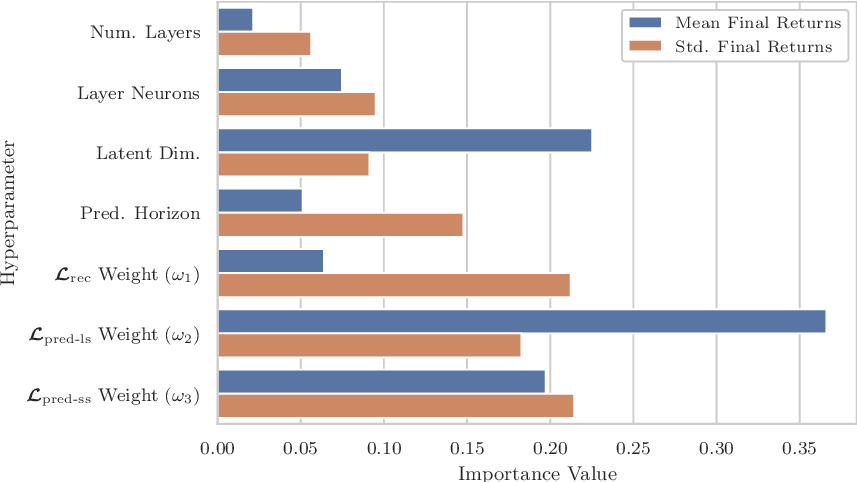}
  \caption{
    Hyperparameter importance scores derived from a random forest regressor.
  }\label{fig:res_hypers_importance}
}

\title{KIPPO: Koopman-Inspired Proximal Policy Optimization}

\author{
Andrei Cozma\(^1\)
\and
Landon Harris\(^1\)
\and
Hairong Qi\(^1\)\\
\affiliations
\(^1\)University of Tennessee, Knoxville\\
\emails
acozma@vols.utk.edu,
lharri73@vols.utk.edu,
hqi@utk.edu
}

\begin{document}

\maketitle


\begin{abstract}

  \gls{RL} has made significant strides in various domains, and policy gradient methods like \gls{PPO} have gained popularity due to their balance in performance, training stability, and computational efficiency.
  These methods directly optimize policies through gradient-based updates.
  However, developing effective control policies for environments with complex and non-linear dynamics remains a challenge.
  High variance in gradient estimates and non-convex optimization landscapes often lead to unstable learning trajectories.
  Koopman Operator Theory has emerged as a powerful framework for studying non-linear systems through an infinite-dimensional linear operator that acts on a higher-dimensional space of measurement functions.
  In contrast with their non-linear counterparts, linear systems are simpler, more predictable, and easier to analyze.
  In this paper, we present \gls{KIPPO}, which learns an approximately linear latent-space representation of the underlying system's dynamics while retaining essential features for effective policy learning.
  This is achieved through a Koopman-approximation auxiliary network that can be added to the baseline policy optimization algorithms without altering the architecture of the core policy or value function.
  Extensive experimental results demonstrate consistent improvements over the \gls{PPO} baseline with 6--60\% increased performance while reducing variability by up to 91\% when evaluated on various continuous control tasks.

\end{abstract}

\section{Introduction}
\label{sec:intro}

\gls{RL} provides a powerful framework for sequential decision-making tasks, enabling agents to learn optimal behaviors through interaction with their environment \citep{sutton2018reinforcement}.
Policy optimization, a core component of this framework, determines the optimal mapping from states to actions that maximizes an agent's cumulative returns.
Policy gradient methods excel in continuous control tasks by directly optimizing policies through gradient-based updates \citep{sutton1999policy}.
However, developing effective control policies for environments with complex and non-linear dynamics remains a challenge.
This challenge, combined with non-convex optimization landscapes, leads to high-variance gradient estimates and unstable updates.
The optimization process often diverges or oscillates, impeding convergence to optimal policies.

The field of dynamical systems studies mathematical models of evolving processes, focusing on patterns, stability, and long-term behavior \citep{broer2010dynamical,heij2021introduction}.
Although linear systems are more predictable, many real-world systems are non-linear, where small changes in initial conditions can lead to drastically different outcomes \citep{mezic2004uncertainty}.
Koopman Operator Theory, a powerful tool for studying non-linear systems, finds a linearized description in a higher-dimensional space of measurement functions, known as the Koopman observable space \citep{koopman1931hamiltonian,brunton2021modernkoopman}.
This process maps original state variables to observable functions, extracting useful state information.
The Koopman operator, an infinite-dimensional linear operator, evolves these observables linearly in time, enabling linear descriptions of non-linear systems.
Data-driven methods like \gls{DMD} and deep learning advances have enabled approximating the Koopman operator directly from data \citep{schmid2010dynamic,yeung2017learning,lusch2018deep}.

\begin{figure*}[tbh]
  \figResultsEnvsMain{}
\end{figure*}

Building on these foundations, we propose \gls{KIPPO}, a method that uses Koopman-inspired representation learning to address a key challenge of policy gradient methods like \gls{PPO}: high-variance gradient estimates in complex, non-linear environments.
Rather than seeking perfectly linear representations of non-linear systems, our approach introduces an inductive bias that encourages approximate linearity along policy trajectories.
This soft constraint simplifies underlying dynamics while preserving essential features for policy learning.
We achieve this through a Koopman-approximation auxiliary network and targeted constraints that balance the complexity of latent dynamics.
\gls{KIPPO}'s architecture uses state encoders/decoders and linear transition matrices to predict future states over a fixed horizon, imposing structure on the latent space while minimizing information loss.
By combining advances in deep learning with Koopman theory principles, this approach simplifies system behavior and improves policy performance.

Our approach creates a mutually beneficial feedback loop: policy gradients identify important state space regions through exploratory rollouts, while our linearization technique reduces gradient variance specifically in these critical regions.
By focusing linearization efforts locally along policy-explored trajectories instead of attempting global linearization.
This targeted approach maintains computational efficiency while delivering benefits precisely where they matter most for the current policy.

The main contribution of this study is threefold:
\begin{enumerate}

  \item \textbf{Koopman-Inspired Policy Optimization.}
        We propose \gls{KIPPO}, an on-policy algorithm that incorporates Koopman operator principles directly into policy gradient updates.
        By learning an approximately linear latent-space representation, \gls{KIPPO} stabilizes gradient estimates and enhances control over non-linear dynamics.

  \item \textbf{Decoupled Auxiliary Representation Learning.}
        \gls{KIPPO} adds an auxiliary network to policy gradient baselines like \gls{PPO} without altering the core policy or value function architecture.
        This design allows the policy to train on a simpler, encoded state space while the auxiliary network enforces a linear-like structure.
        As a result, standard \gls{PPO} hyperparameters and training loops remain largely intact.

  \item \textbf{Performance and Stability Improvements.}
        Across MuJoCo and Box2D tasks, \gls{KIPPO} consistently achieves 6--60\% higher mean returns and a 26--91\% reduction in variance compared to baseline \gls{PPO}, as shown in Fig.~\ref{fig:results_envs_main_percdiff}.
        These empirical gains attest to the efficacy of Koopman-inspired constraints in mitigating high-variance updates and accelerating convergence.

\end{enumerate}

The remainder of this paper is organized as follows:
Sec.~\ref{sec:background} presents essential background.
Sec.~\ref{sec:method} describes the \gls{KIPPO} framework in detail.
Sec.~\ref{sec:exp_res} presents extensive experimental results across multiple environments.
Sec.~\ref{sec:conclusion} concludes the paper by summarizing our findings and outlining promising directions for future research.

\section{Background and Related Works}\label{sec:background}

\gls{RL} provides a framework for agents to learn optimal behaviors through interactions with their environment.
These interactions are formalized as a \gls{MDP}, defined by a tuple \((\gls{state-space}, \gls{action-space}, \mathcal{P}, \gls{reward-func}, \gls{discount})\), where \(\gls{state-space}\) represents the state space, \(\gls{action-space}\) the action space, \(\mathcal{P}: \gls{state-space} \times \gls{action-space} \times \gls{state-space} \to [0, 1]\) the transition probability function, \(\gls{reward-func}: \gls{state-space} \times \gls{action-space} \to \gls{reals}\) the reward function, and \(\gls{discount} \in [0, 1]\) the discount factor.

This interaction mirrors the feedback loop in control systems, where the agent acts as the controller and the environment represents the system being controlled.
The environment in \gls{RL} can then be modeled as a continuous-time dynamical system:
%
$ \gls{state}_{t+1} = \gls{dyn-trans-func}(\gls{state}_t, \gls{action}_t)$,
%
where \(\gls{state}_t \in \gls{state-space}\) and \(\gls{action}_t \in \gls{action-space}\) are the state and action, respectively, at time \(t\), and \(\gls{dyn-trans-func}: \gls{state-space} \times \gls{action-space} \to \gls{state-space}\) is the (often non-linear) state transition function.

\textbf{\gls{RL} and Policy Gradient Methods.} \gls{RL} algorithms fall into several categories: {\it value-based methods}
and {\it policy-based methods}.
The latter can be further categorized into {\it on-policy methods} that learn exclusively from current policy experiences and {\it off-policy methods} that can learn from any policy's experiences.
\gls{RL} algorithms can also be grouped by model-based versus model-free approaches, distinguished by whether they explicitly learn environment dynamics.

Generally speaking, policy optimization algorithms learn a parameterized policy \(\gls{policy}_{\gls{params}}\) by optimizing parameters \(\gls{params}\) through gradient descent with respect to the expected return.
While successful in various tasks, policy gradient methods face fundamental challenges with complex, non-linear dynamics.
High variance makes gradient estimates less reliable, and non-convex optimization landscapes often lead to unstable learning trajectories.

Actor-critic methods, such as \gls{A2C} \citep{mnih2016asynchronous}, address these issues by using a value function (critic) to provide lower-variance targets for policy (actor) updates.
Trust region methods like \gls{TRPO} \citep{schulman2017trustregionpolicyoptimization} constrain policy updates using \gls{KL} divergence to ensure new policies remain within a trusted region.
However, \gls{TRPO}['s] second-order optimization approach increases computational complexity.
\gls{PPO} \citep{schulman2017proximalpolicyoptimizationalgorithms} addresses these limitations with a first-order approach that avoids Hessian computations while maintaining trust region properties, becoming a leading algorithm for continuous and discrete control tasks.
For its policy component, it introduces a clipped surrogate objective function:
$
  L^{CLIP}(\theta) = \hat{\mathbb{E}}_t\left[\min\left(r_t(\theta)\hat{A}_t, \text{clip}(r_t(\theta), 1-\epsilon, 1+\epsilon)\hat{A}_t\right)\right]
$
where $\hat{\mathbb{E}}_t$ denotes the empirical average over timesteps, $r_t(\theta) = \frac{\pi_\theta(a_t|s_t)}{\pi_{\theta_{old}}(a_t|s_t)}$ is the probability ratio between policies, $\hat{A}_t$ is the estimated advantage, and $\epsilon$ is a hyperparameter limiting policy change (typically 0.2 for continuous domains, 0.1 for discrete tasks).

The algorithm alternates between collecting experiences and optimizing a combined objective (clipped policy, value function, and entropy terms) via minibatch gradient ascent.
This achieves the same policy constraints as \gls{TRPO} without second-order derivatives, improving performance while maintaining efficiency.

Recent advances include \gls{SLAC} \citep{lee2020stochastic}, which integrates policy optimization with latent state representation learning using variational inference for improved sample efficiency;
\gls{MBPO} \citep{janner2021trustmodelmodelbasedpolicy}, which employs ensemble dynamics models for synthetic experience generation;
and \gls{RPO} \citep{rahman2022robustpolicyoptimizationdeep}, which extends \gls{PPO} with perturbed action distributions to maintain policy entropy and enhance robustness.
Nevertheless, achieving robust generalization and efficient learning in complex, non-linear systems remains an open challenge.

\textbf{Koopman Operator Theory} provides a mathematical framework to transform non-linear dynamics into linear operators acting on observable functions \citep{koopman1931hamiltonian,mezic2005spectral,rowley2009spectral}.
The key insight is that while system dynamics may be highly non-linear in state space, they can be represented linearly in an appropriate space of observable functions (measurement functions that extract system information).
For systems with control inputs, the Koopman formulation is:
\begin{equation}
  \gls{state-obs-func}(\gls{state}_{t+1}) = \gls{state-obs}_{t+1} = \gls{koop-mat-state} \gls{state-obs}_t + \gls{koop-mat-action} \gls{action-obs}_t
  \label{eq:koopman_control}
\end{equation}
where $\gls{state-obs}_t = \gls{state-obs-func}(\gls{state}_t) \in \gls{reals}^m$ are state observables, $\gls{action-obs}_t = \gls{action-obs-func}(\gls{action}_t) \in \gls{reals}^k$ are control observables, and $\gls{koop-mat-state} \in \gls{reals}^{m\times m}$ and $\gls{koop-mat-action} \in \gls{reals}^{m \times k}$ are finite-dimensional matrices approximating the infinite-dimensional Koopman operator.
The challenge lies in finding appropriate observable functions that enable effective linearization, which can be learned using deep neural networks \citep{lusch2018deep,dey2023dlkoopman}.

\textbf{Research integrating Koopman theory with control and \gls{RL}} has progressed along two paths: system modeling with control, and integration with model-free \gls{RL} algorithms.
The former includes work by \citet{han2020deep} and \citet{shi2022deep}, who developed frameworks combining Koopman operators with linear control methods like \glspl{LQR} and \glspl{MPC}.
\citet{yin2022embedding} combined Koopman theory with \gls{LQR} to create differentiable policies embedding optimal control principles.
The latter focuses on model-free \gls{RL}, including work by \citet{song2021data} that introduced \gls{DKRL}, which uses local Koopman operators to improve data efficiency.
\citet{weissenbacher2022koopman} proposed \gls{KFC}, an offline algorithm that leverages Koopman theory to infer symmetries in system dynamics.

While model-based methods using Koopman theory with \glspl{MPC} have been thoroughly investigated \citep{korda2018linear}, they typically incur high computational costs due to optimization requirements at each timestep.
Model-free approaches avoid this overhead but have received less attention.

\gls{KIPPO} differs from existing approaches by 1) focusing on on-policy learning improvement rather than global linearization, 2) integrating representation learning directly into policy optimization, 3) decoupling representation learning from policy updates, and 4) measuring success through policy performance and stability rather than global approximation quality.

\section{Methodology}\label{sec:method}

\begin{figure*}[htb]
  \figArchitecture{}
\end{figure*}

\gls{KIPPO} introduces a Koopman-inspired representation learning framework that operates independently alongside the core policy optimization process.
This approach unifies traditional \gls{RL} with Koopman operator theory while maintaining practical implementability.
The framework builds on several key principles:

\begin{itemize}
  \item \textbf{Decoupled Optimization}: Representation learning and policy optimization are deliberately separated to prevent interference between objectives, where the core policy algorithm remains unchanged, operating on encoded states without modification to its optimization process
  \item \textbf{Local Linearization}: Rather than attempting global linearization, the framework focuses on simplifying dynamics along policy-explored trajectories
  \item \textbf{Balanced Complexity}: Loss functions are designed to balance the competing objectives of simplification and information preservation
\end{itemize}

\gls{KIPPO}['s] key innovation is its targeted approach to linearity, expressed as $\gls{state-encoder}(\gls{state}_{t+1}) \approx \gls{koop-mat-state} \gls{state-encoder}(\gls{state}_t) + \gls{koop-mat-action} \gls{action-encoder}(\gls{action}_t)$.
Unlike networks with standard linear output layers, \gls{KIPPO} enforces linear dynamics across time steps as a soft constraint, applying this only to policy-explored trajectories.
This creates an inductive bias on temporal transitions rather than static mappings, promoting stable gradient flow while avoiding the computational burden of global linearization.

\subsection{Architecture Design}\label{sec:architecture}

Building upon these design principles, the core innovation of \gls{KIPPO} lies in learning a latent representation where complex, non-linear environment dynamics can be effectively approximated by linear operations within regions of the state space explored by the current policy.

While sharing similarities with representation learning and data compression, \gls{KIPPO} employs a higher-dimensional latent space than the original state space.
This design choice follows from Koopman theory, which demonstrates that non-linear dynamics can be linearized through appropriate lifting to higher-dimensional spaces of observable functions \citep{budisic2012applied}.

As shown in Fig.~\ref{fig:kippo-architecture}, \gls{KIPPO} comprises several interconnected components:

\begin{itemize}
  \item A state autoencoder consisting of an encoder \(\gls{state-encoder}: \gls{state-space} \to \gls{reals}^m\) and decoder \(\gls{state-decoder}: \gls{reals}^m \to \gls{state-space}\), which respectively map states to latent representations and reconstruct original states
  \item An action encoder \(\gls{action-encoder}: \gls{action-space} \to \gls{reals}^k\) that maps actions to the latent space
  \item Linear system matrices \(\gls{koop-mat-state} \in \gls{reals}^{m\times m}\) and \(\gls{koop-mat-action} \in \gls{reals}^{m\times k}\) that govern the dynamics within the latent space
\end{itemize}

The encoder and decoder networks use \glspl{MLP} with hyperbolic tangent (tanh) activation functions, chosen for their smooth gradients and bounded output range.
All trainable parameters employ Xavier uniform initialization \citep{glorot2010understanding} to promote stable learning in deep networks, except for \(\gls{koop-mat-state}\), which uses orthogonal initialization \citep{saxe2014exact} for stable gradient flow, and \(\gls{koop-mat-action}\), which starts with zeros to allow gradual learning of control effects.
The number of layers and neurons per layer remain consistent across the state encoder, decoder, and action encoder networks, typically using 2-3 hidden layers with 64-256 units each.
This architectural consistency helps maintain balanced representational capacity across components while remaining computationally efficient.

The use of non-linear activation functions might appear counterintuitive given our goal of linear dynamics.
However, these non-linearities are essential for learning effective Koopman observables, enabling the networks to discover appropriate lifting functions that map the original system to a space where linear approximations become effective along policy-relevant trajectories.
While the Koopman operator governing the evolution of observables is inherently linear, the method of obtaining these observables need not be linear.
These non-linearities enable the \glspl{MLP} to act as universal function approximators, making them well-suited for approximating the Koopman observables in the latent space.

The dimensions of the state-transition matrix \(\gls{koop-mat-state}\) and control matrix \(\gls{koop-mat-action}\) correspond to the chosen latent space dimensionality.
This is typically set to 2-4 times the state dimension, providing sufficient capacity to capture complex dynamics without excessive computational overhead.
These matrices are learnable parameters optimized alongside other components, enabling the framework to learn environment-specific representations directly from experience.

These components work together to approximate the Koopman operator's action on observable functions, with the encoders serving as learnable observable functions and the linear matrices capturing the evolution of these observables.
This connection to Koopman theory provides theoretical grounding for our approach while remaining practically implementable within the \gls{RL} context.

\subsection{Future State Prediction Process}\label{sec:prediction}

The prediction process forecasts states over horizon \(\gls{horizon}\) using learned linear latent-space dynamics.
For an initial state \(\gls{state}_0\) and an action sequence \(\gls{action}_{0:\gls{horizon}}\), the process begins with the initial encoding of the state into a latent representation, \(y_0 = \phi_x(x_0)\).
This is followed by an iterative prediction using learned dynamics, expressed as \(\gls{state-pred-obs}_{h+1} = \gls{koop-mat-state} y_{h} + \gls{koop-mat-action} \gls{action-encoder}(\gls{action}_{h})\).
This process yields a sequence of predicted latent states, which can be decoded back to the original state space using \(\gls{state-pred}_{h+1} = \gls{state-decoder}(\gls{state-pred-obs}_{h+1})\).

The process implements the finite-dimensional approximation of the Koopman-based control formulation from Eq.~\ref{eq:koopman_control}, where
\(\gls{state-encoder}\) and \(\gls{action-encoder}\) represent \(\gls{state-obs-func}\) and \(\gls{action-obs-func}\), respectively.
This process primarily constrains the learning of the latent representation to reduce gradient variance, rather than for generating additional training data or performing planning.
Importantly, this serves as a soft constraint; perfect linearity is not required, but the representation is encouraged to be approximately linear along policy trajectories to enable more stable policy optimization.

The multi-step prediction shapes representations by enforcing temporal consistency only along current-policy trajectories, avoiding unrealistic global linearity assumptions.
This allows \gls{KIPPO} to benefit from structured representations without requiring lookahead or model predictive control.
Unlike model-based planning methods, we never use the learned model for planning; our method focuses on variance reduction through temporal coherence.
We deliberately chose this novel application of predictive models solely for variance reduction rather than for planning.
It specifically addresses the noisy updates that challenge policy gradient methods.

The prediction horizon \(\gls{horizon}\) balances computational cost against prediction depth.
Empirically, horizons of 8-32 steps are effective, with longer horizons benefiting environments with significant temporal dependencies or sparse rewards.
As analyzed in Appendix~\ref{app:hypers:horizon} and Table~\ref{tab:effect_horizon}, longer horizons benefit moderately complex environments but offer diminishing returns or instability in very simple or highly complex ones.

Detailed implementation steps for the prediction process are provided in the supplementary material (\Cref{app:impl:pred}).

\subsection{Loss Formulation}\label{sec:losses}

Drawing inspiration from Koopman operator theory, \gls{KIPPO} employs three complementary loss components, reconstruction loss, latent-space prediction loss, and state-space prediction loss, that shape the latent space to achieve four key properties, including 1) \textit{informativeness} where essential information from the original state space is preserved, ensuring the agent can make decisions based on accurate representations, 2) \textit{simplification} where system dynamics is represented using linear approximation specifically along policy trajectories, rather than globally across the entire state space, 3) \textit{predictability} where accurate multi-step predictions can be achieved within explored regions, enabling better temporal coherence and reduced gradient variance, and 4) \textit{consistency} where the representation aligns with true environment dynamics for effective policy learning.

\subsubsection{Reconstruction Loss}\label{subsec:reconstruction_loss}

The reconstruction loss ensures the latent space retains sufficient information for accurate state reconstruction:

\begin{equation}
  \gls{loss-ki-rec}(t) = \left\{\gls{state-decoder}(\gls{state-encoder}(\gls{state}_t)) - \gls{state}_t\right\}^2
  \label{eq:loss_rec}
\end{equation}

This loss primarily addresses ``informativeness'' while aligning with Koopman theory principles, where observable functions are typically assumed to be invertible.
This formulation allows a bijective mapping between the original state space and latent space, maintaining a meaningful connection that supports both representation learning and policy optimization.

Note that the action reconstruction loss is omitted since the sole purpose of the action encoder is to influence state transitions in the latent space, and the accuracy of action encoding is implicitly enforced through future state prediction losses.
Empirical studies also confirm that including action reconstruction terms does not yield significant performance improvements.

\subsubsection{Latent-Space Prediction Loss}\label{subsec:latent_prediction_loss}

The latent-space prediction loss primarily targets ``simplification'' and ``predictability'' within policy-relevant regions.
It encourages learning a representation where dynamics can be effectively approximated by linear operations along policy-explored trajectories:

\begin{equation}
  \gls{loss-ki-pred-ls}(t) = \frac{1}{\gls{horizon}} \sum_{h=1}^{\gls{horizon}} \gls{binary-mask}_{t,h}\,\bigl(\gls{state-pred-obs}_{t+h} - \gls{state-encoder}(\gls{state}_{t+h})\bigr)^2
  \label{eq:loss_pred_latent}
\end{equation}
where \(\gls{binary-mask}_{t,h}\) handles episode boundaries through a binary mask:
\begin{equation}
  \gls{binary-mask}_{t,h} =
  \begin{cases}
    1, & \text{if trajectory not ended by step }(t + h - 1), \\
    0, & \text{otherwise.}
  \end{cases}
  \label{eq:binary_mask}
\end{equation}

The binary mask is essential for handling variable-length trajectories, such that the prediction process is not penalized for discontinuities introduced by environment resets at episode boundaries.

The loss term facilitates the learning of representations where dynamics can be effectively approximated by linear operations, as \(\gls{state-pred-obs}_{t+h}\) is generated using linear matrices \(\gls{koop-mat-state}\) and \(\gls{koop-mat-action}\).
It also enhances predictability by minimizing multi-step prediction errors directly in the latent space, while supporting simplification by encouraging the encoder to find representations where linear predictions maintain accuracy over multiple timesteps.

This loss term is particularly important for maintaining the framework's Koopman-inspired aspects, as it drives the learning of representations that align with Koopman theory's principle of lifting non-linear dynamics to spaces where linear approximations become effective.

\subsubsection{State-Space Prediction Loss}\label{subsec:state_prediction_loss}

The state-space prediction loss primarily addresses ``consistency'' and ``predictability'', maintaining fidelity to true dynamics while ensuring meaningful state predictions:

\begin{equation}
  \gls{loss-ki-pred-ss}(t) = \frac{1}{\gls{horizon}} \sum_{h=1}^{\gls{horizon}} \gls{binary-mask}_{t,h}\,\bigl(\gls{state-decoder}(\gls{state-pred-obs}_{t+h}) - \gls{state}_{t+h}\bigr)^2
  \label{eq:loss_pred_state}
\end{equation}

The loss helps prevent the latent space from diverging too far from physically meaningful representations, which is essential for learning effective control policies.

This dual-space prediction approach ensures the latent dynamics align with true environment behavior when mapped back to state space.
It also promotes learning of latent representations that maintain predictive power across multiple timesteps, indirectly reinforcing informativeness by requiring accurate long-term state reconstruction.
The improved temporal consistency from these representations helps reduce gradient variance in policy updates, though we never use these predictions for planning.

\subsubsection{Total Representation Loss}\label{subsec:total_loss}

The total representation loss is a weighted sum of the three components:

\begin{equation}
  \gls{loss-ki-wsum} = \frac{1}{\gls{total-steps}}\sum_{t=0}^{\gls{total-steps}} \left(\gls{loss-weight-ki-rec} \gls{loss-ki-rec}(t) + \gls{loss-weight-ki-pred-ls} \gls{loss-ki-pred-ls}(t) + \gls{loss-weight-ki-pred-ss} \gls{loss-ki-pred-ss}(t)\right)
  \label{eq:loss-ki-wsum}
\end{equation}
where \(\gls{total-steps}\) represents the number of steps collected during rollouts.
The weights \(\gls{loss-weight-ki-rec}\), \(\gls{loss-weight-ki-pred-ls}\), and \(\gls{loss-weight-ki-pred-ss}\) incorporate several factors, including relative scales between latent and state space dimensionalities, task-specific requirements, and environment characteristics.
Sec.~\ref{sec:ablation} investigates the impact of each loss term on both the return and stability of learning.

\subsection{Overview of Training Process}\label{sec:training}

The training process in \gls{KIPPO} alternates between rollout and optimization phases until reaching a predetermined number of environment steps.
During rollouts, the agent collects states, actions, rewards, and additional sequences needed for representation losses, storing them in separate buffers.
Each rollout phase collects 2,048 environment steps across multiple trajectories, resetting the environment when necessary to ensure diverse experiences.

Both \gls{KIPPO} and \gls{PPO} use identical on-policy rollouts and operate with the same available information.
\gls{KIPPO} utilizes future states solely as auxiliary loss targets (never as policy inputs), which is consistent with standard auxiliary objective practices in on-policy \gls{RL}.
During both training and inference, both methods receive identical trajectories and current-state information, with \gls{KIPPO} applying state encoding while \gls{PPO} uses raw states.

The optimization phase processes the collected data to update all components.
The algorithm divides 2,048 steps into 32 mini-batches, computing the three key losses to update representation learning components.

The total framework loss combines the weighted representation loss and standard \gls{PPO} loss:

\begin{equation}
  \gls{loss-kippo} = \gls{loss-ki-wsum} + \gls{loss-ppo}
  \label{eq:loss-kippo}
\end{equation}
Both components update their parameters using the Adam optimizer.
The optimization process runs for 10 epochs, allowing refinement of both the latent representation and the policy.
After optimization, a new rollout phase begins, continuing this cycle until reaching 1 million environment steps.

The latent representation is learned incrementally throughout training, with parameters adapting gradually across rollout-optimization cycles.
We observe stability, with representations evolving smoothly between updates, maintaining consistent state encodings, and preventing disruptive changes that could destabilize learning.
This is particularly important for policy gradient methods sensitive to sudden representation shifts.

A key feature is the complete decoupling of representation learning from policy optimization.
The representation learning components
optimize independently from policy and value networks.
This separation ensures improvements stem from the learned representation.
Detailed optimization phase implementation and pseudocode are provided in Appendix~\ref{app:impl:optim}.

\section{Experiments and Results}\label{sec:exp_res}

We evaluate the effectiveness of \gls{KIPPO} compared to baseline \gls{PPO} and \gls{RPO} algorithms across diverse continuous control tasks, measuring both performance improvements and reduced variability.

\subsection{Experimental Setup}\label{sec:setup}

\subsubsection{Environments}\label{sec:setup:envs}

We evaluate six continuous control environments from Gymnasium \citep{towers2023gymnasium} using MuJoCo \citep{todorov2012mujoco} and Box2D \citep{box2d}, forming a comprehensive testbed with diverse complexity levels and control challenges.
The environments' varying non-linearity and temporal dependencies help evaluate the algorithm's robustness and its ability to learn effective representations in the Koopman observable space.
We chose these testbeds to systematically evaluate how our approach reduces gradient variance across different complexity levels while keeping the analysis tractable.

To facilitate discussion, we classify the six environments by their complexity levels, defined using the dimensions of the state space, $|\mathcal{S}|$, and the action space, $|\mathcal{A}|$.
An environment is considered to have ``low'' complexity if $|\mathcal{S}| + |\mathcal{A}| < 10$, ``medium'' complexity if $ 10 \leq |\mathcal{S}| + |\mathcal{A}| < 20$, and ``high'' complexity if $|\mathcal{S}| + |\mathcal{A}| \geq 20$.
This is summarized in \Cref{tab:envs_difficulty}.
For detailed specifications of the environments, please refer to \Cref{app:setup:envs}.

\begin{table}[htb]
  \tabEnvsDifficulty{}
\end{table}

\subsubsection{Training Configuration}\label{sec:setup:config}

For meaningful comparisons, we implement our benchmarks using the \gls{PPO} and \gls{RPO} implementations from the CleanRL library \citep{huang2022cleanrl}.
We maintain CleanRL's default hyperparameters for both algorithms.

Each experiment uses 4 random initialization seeds (1, 2, 3, 4) per environment.
We selected 4 seeds as a balance between the original \gls{PPO} paper's 3 seeds \citep{schulman2017proximalpolicyoptimizationalgorithms} and CleanRL's standard 5 seeds.
These seeds determine both the environment's initial states and model parameter initialization.
To ensure fair comparison, we use identical random seeds and initialization patterns across all methods.
Each training run consists of exactly 1 million environment steps.
We provide hardware specifications and runtime measurements in Appendix~\ref{app:setup:compute}.

\subsubsection{Evaluation Metrics}\label{sec:setup:metrics}

Given the inherent stochasticity in both environment and learning processes, we employ the \gls{EWMA} of episodic returns to capture learning trends effectively:
\begin{equation}
  \text{\glsentryshort{EWMA}}_t = \alpha \cdot \text{\glsentryshort{EWMA}}_{t-1} + (1 - \alpha) \cdot \gls{return}_t,
  \label{eq:metric_ewma}
\end{equation}
where $G_t$ is the expected (discounted) return, summing rewards weighted by $\gamma^t$ at each time step; and empirically determined \(\alpha=0.05\).
$\alpha$ balances responsiveness to recent changes with historical context, reducing noise by filtering short-term fluctuations and ensuring robustness against outliers.

We also use the \gls{CTE} metric:

\begin{equation}
  \text{\glsentryshort{CTE}} = \frac{1}{\gls{horizon}} \sum_{h=1}^{\gls{horizon}} \frac{1}{h}\sum_{k=1}^{h} \lvert \gls{state-pred}_k - \gls{state}_k \rvert
  \label{eq:metric_cte}
\end{equation}
where $k$ indices the individual timestep.
While \gls{EWMA} evaluates the overall agent performance, \gls{CTE} specifically measures the representation quality by comparing predicted states (\gls{state-pred}) with actual states (\gls{state}) across varying horizons.

For a comprehensive evaluation, we analyze both metrics through their means and standard deviations (SD) across 4 independent training runs.

\subsection{Comparison with Baselines}\label{sec:comparison}

We conduct the first set of experiments by comparing \gls{KIPPO}['s] performance with two baseline policy gradient methods, \gls{PPO} and \gls{RPO}, selected due to either popularity or state-of-the-art performance.
The comparison is summarized in \Cref{tab:results_envs_main}.
Fig.~\ref{fig:results_envs_main_percdiff} presents the percent difference in these metrics relative to the \gls{PPO} baseline.
\begin{table*}[tb]
  \tabResultsEnvsMain{}
\end{table*}

From both \Cref{tab:results_envs_main} and Fig.~\ref{fig:results_envs_main_percdiff}, we observe that
\gls{KIPPO} achieves overwhelmingly better mean performance in all environments, with improvements ranging from 6.36\% to 60.26\% for the \gls{PPO} baseline and from 11.86\% to 142.18\% for the \gls{RPO} baseline.
\gls{KIPPO} also shows lower SD in most environments, demonstrating enhanced consistency across seeds, reducing variance by 26.89-91.43\% versus \gls{PPO} (one exception) and 58.94-90.21\% versus \gls{RPO} (two exceptions).
We will further discuss the exceptional cases in the ablation study.

This dual improvement in performance and stability highlights the fundamental advantage of incorporating Koopman-inspired representation.
For a more granular view of performance evolution, we provide detailed learning curves in \Cref{app:train_curves}.

\subsection{Ablation Study of Loss Components}\label{sec:ablation}

To understand the mechanisms underlying \gls{KIPPO}['s] performance advantages, we conduct a systematic ablation study of its core components.
\Cref{tab:effect_losses_partial} shows the result of one environment.
For the other five environments, please refer to \vref{tab:effect_losses} in \Cref{app:ablation_losses}.

\begin{table*}[tb]
  \tabEffectLossesPartial{}
\end{table*}

Several key findings emerge from Figs.~\ref{fig:rel_impr_means_loss_combinations}, \ref{fig:rel_impr_stds_loss_combinations} and Tables~\ref{tab:effect_losses_partial}, \ref{tab:effect_losses}.
Note that the first row per environment represents baseline \gls{PPO} performance.
First, the reconstruction loss alone yields results comparable to those of the baseline.
Second, integrating all losses produces the best mean \gls{EWMA} and lowest standard deviation in most cases.
Third, the mean \gls{CTE} decreases systematically with the incorporation of additional loss components.
And finally, the latent-space prediction loss consistently reduces both \gls{CTE} mean and standard deviation.

While the ablation results in Tables~\ref{tab:effect_losses_partial} and \ref{tab:effect_losses} highlight the contributions of each loss component, they also reveal that \gls{KIPPO}['s] overall gains depend strongly on the level of non-linearity (complexity) of each environment.
We hypothesize a \emph{sublinear} (``logarithmic'') relationship between performance gain and complexity,
meaning that beyond a certain point, additional non-linearity or complexity diminishes marginal returns and raises variance.

\begin{figure}[htbp]
  \centering
  \includegraphics[width=1.0\linewidth]{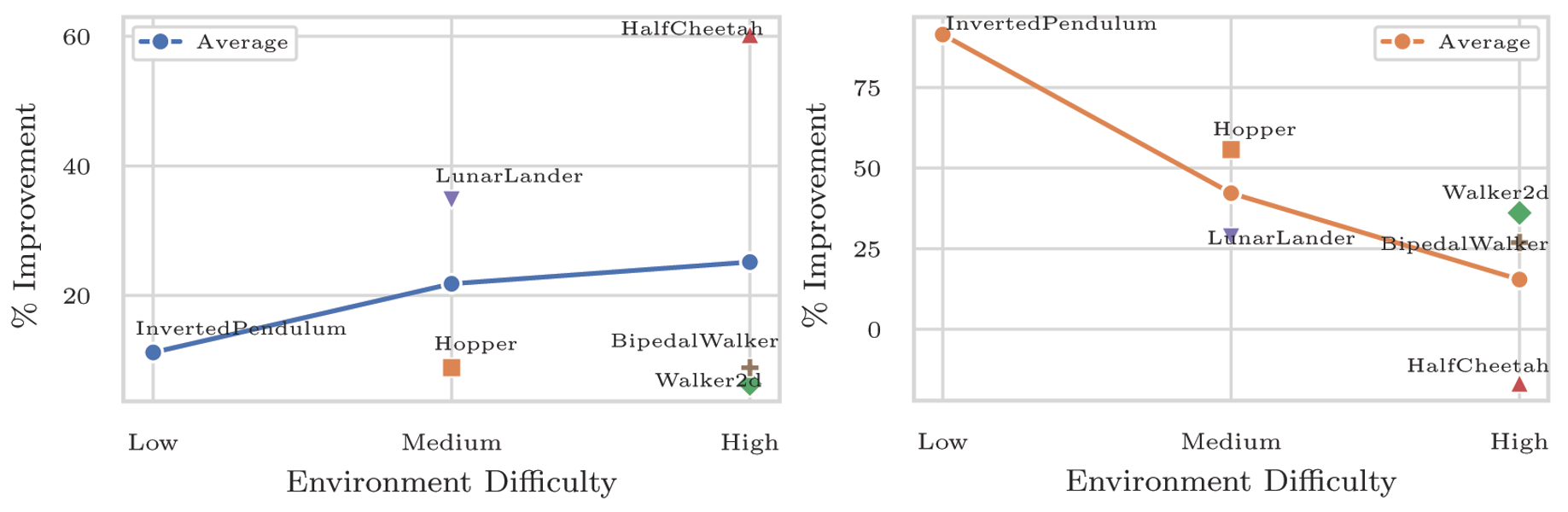}
  \caption{
    The mean percent improvement of environments with various levels of complexity in performance gain (Left) and variance reduction (Right) of final returns by \glsentryshort{KIPPO} compared to \glsentryshort{PPO}.
  }
  \label{fig:rel_impr_means_combined}
\end{figure}

Fig.~\ref{fig:rel_impr_means_combined} shows our evaluation of \glsentryshort*{KIPPO} across varying environment complexities.
We analyze two key metrics compared to the \glsentryshort*{PPO} baseline: relative improvement in average performance (mean) and consistency of results (SD).

Overall, \gls{KIPPO} scales effectively across a broad range of non-linear control tasks but hits an upper limit as complexity grows.
In simpler domains, overemphasizing latent-space predictions can harm stability unless balanced by state-space constraints.
In extreme tasks, significant raw gains come with heightened variance.
Thus, \gls{KIPPO} extends \gls{PPO}'s performance boundary significantly,
but the underlying non-linearities impose a \emph{logarithmic} or sublinear bound on further improvements.

\subsection{Sensitivity Analysis and Limitations}\label{sec:sensitivity}

In this set of experiments, we conduct extensive parameter sensitivity studies, particularly focusing on the latent dimension and prediction horizon to better understand \gls{KIPPO}['s] limitations.
Comprehensive quantitative results are available in Appendices~\ref{app:hypers:latent} and~\ref{app:hypers:horizon}.

We observe that performance gains diminish in environments with highly discontinuous transitions (e.g., collisions), contact-rich interactions, or multi-modal behaviors, as the linear latent dynamics struggle with abrupt changes.
Despite enforcing approximate linearity only along policy trajectories as a soft constraint, environments with highly chaotic dynamics remain challenging.

In environments with sparse rewards, the advantage over baseline \gls{PPO} is less pronounced, suggesting that reduced gradient variance benefits are most impactful with frequent feedback signals.
We view Koopman-based dynamics as an inductive bias particularly well-suited for certain control problems rather than as a universally valid model.
These selected environments provide a controlled setting to test our core hypothesis: linearized latent dynamics can reduce gradient variance in policy optimization.

Training \gls{KIPPO} takes approximately 15\% longer than \gls{PPO} (15 hours vs. 13 hours for 24 parallel models) due to (1) construction of prediction sequences and (2) computation of multi-step prediction losses.
However, this computational overhead exists only during training; at inference time, only the encoder is used with negligible additional computational cost.

To identify the most influential hyperparameters, we train a random forest regressor to predict final returns from hyperparameter configurations.
Fig.~\ref{fig:res_hypers_importance} shows that the latent-space prediction loss weight (\gls{loss-weight-ki-pred-ls}) has the highest importance, followed by the latent dimension and state-space prediction loss weight (each ~0.20).
For return variability, the three loss weights (each ~0.20) dominate, followed by the prediction horizon (0.15).

\begin{figure}[htb]
  \figHypersImportance{}
\end{figure}

\section{Conclusions and Future Work}\label{sec:conclusion}

\gls{KIPPO} addresses key challenges in policy gradient methods through stable policy optimization for complex non-linear control tasks.
Our experiments demonstrate the effectiveness of Koopman-inspired representation learning in policy optimization as showcased in \gls{PPO} and \gls{RPO}.
This architecture naturally extends to other on-policy algorithms, including \gls{TRPO} and \gls{A2C}.

The effectiveness of \gls{KIPPO} stems from a synergistic bidirectional relationship: policy gradients generate exploratory rollouts that guide which latent regions to linearize, while the resulting representations reduce gradient variance in precisely those regions, creating a more effective feedback loop than decoupled representation learning.
This mechanism retains gradient variance reduction benefits even when extended beyond on-policy methods.

Future directions include extending to: 1) off-policy algorithms like \gls{DDPG}, \gls{TD3}, and \gls{SAC}; 2) value-based methods for enhancing Q-function learning; and 3) discrete domains through appropriate latent space formulations.
Further research opportunities involve handling discontinuous dynamics and investigating representation robustness under noise and distribution shifts.

\bibliographystyle{named}
\bibliography{
  ijcai25_main
}

\appendix

\section{Appendix / Supplementary Material}
\counterwithin{figure}{section}
\counterwithin{table}{section}
\setcounter{table}{0}
\setcounter{figure}{0}


\section{Experimental Setup Details}\label{app:setup}

\subsection{Environment Details}\label{app:setup:envs}

The experiments utilize six environments from the Gymnasium library \citep{towers2023gymnasium}, implemented with MuJoCo \citep{todorov2012mujoco} and Box2D \citep{box2d} physics engines.
These environments represent diverse complexity levels and control challenges:

\begin{enumerate}
  \item \textbf{Inverted Pendulum:}
        Balances a pole on a cart by applying horizontal forces, offering a simple testbed with straightforward dynamics.

  \item \textbf{Lunar Lander:}
        Controls main and side thrusters for precise landing, balancing fuel consumption and landing accuracy.

  \item \textbf{Bipedal Walker:}
        Features uneven terrain and LiDAR input, requiring coordination of multiple joints for balanced forward movement.

  \item \textbf{Hopper:}
        Involves single-leg hopping dynamics, requiring coordinated control for stable movement and forward progression.

  \item \textbf{Walker:}
        Presents a bipedal locomotion task requiring balance and coordination of multiple joints for forward movement.

  \item \textbf{Half Cheetah:}
        Comprises a high-speed locomotion task with a planar body and two legs, focusing on dynamic gaits and efficiency.
\end{enumerate}

These environments were selected to provide a comprehensive evaluation across different levels of difficulty, with diverse reward structures ranging from dense (e.g., Inverted Pendulum) to sparse (e.g., Lunar Lander).
Visual representations of all environments are shown in~\vref{fig:envs_thumbs_two_by_three}.

\begin{figure}[htb]
  \figEnvsThumbsTwoByThree{}
\end{figure}

\subsection{Training Configuration}\label{app:setup:train}

This section details the hyperparameter configurations for baseline models and our approach.
\Vref{tab:ppo_hypers} specifies the fixed hyperparameters for baseline \gls{PPO} and \gls{RPO} models, including optimization, rollout, and policy update settings.

\begin{table}[htb]
  \tabPPOhypers{}
\end{table}

\Vref{tab:kippo_hypers} presents the additional hyperparameters for \gls{KIPPO}, including latent space dimension, prediction horizon, network architecture, and loss weights.
While the latent dimension and prediction horizon varied per environment, other parameters remained fixed.
"AE" denotes choices related to the state auto-encoder, encompassing both the encoder and decoder, as well as the action encoder.

\begin{table}[htb]
  \tabKIPPOhypers{}
\end{table}

\subsection{Hardware and Runtime Details}\label{app:setup:compute}

We conducted experiments in parallel across four seeds and six environments, totaling 24 simultaneous training runs.
Each experiment used a dedicated core of an Intel Xeon Gold 6248R CPU (3.00GHz, 24 cores per socket, 2 sockets).
The system used a single NVIDIA Tesla V100S-PCIe-32GB GPU for one complete set of 24 runs.
This configuration required approximately 15 hours for \gls{KIPPO} and 13 hours for baseline \gls{PPO}, demonstrating practical applicability with modest computational overhead.

\section{Training Curves}\label{app:train_curves}

\Vref{fig:results_envs_main} compares the training curves for \gls{KIPPO} and the \gls{PPO} baseline across the six environments, revealing several notable trends:

\begin{itemize}
  \item \textbf{InvertedPendulum}: The proposed method surpasses the baseline around 0.2 million steps, maintaining superior performance and lower standard deviation thereafter.

  \item \textbf{LunarLander}: While the baseline converges to a return of ~200 midway, our approach continues improving throughout, with similar variability.

  \item \textbf{BipedalWalker}: The proposed method maintains lower variability throughout and surpasses the baseline towards the end, while \gls{PPO} shows increased standard deviation between 0.2-0.5 million steps.

  \item \textbf{Hopper}: The algorithm shows consistent improvement in mean returns, with variability similar to the baseline.

  \item \textbf{Walker2d}: Our method maintains lower variability and surpasses the baseline towards the end, suggesting potential for further improvement with extended training.

  \item \textbf{HalfCheetah}: The proposed approach demonstrates significantly improved mean returns but slightly higher variability compared to the baseline.
\end{itemize}

\begin{figure*}[p]
  \figTrainingCurves{}
\end{figure*}

\section{Ablation Study of Loss Components}\label{app:ablation_losses}

\begin{figure}[tb]
  \centering
  \includegraphics[width=1.0\linewidth]{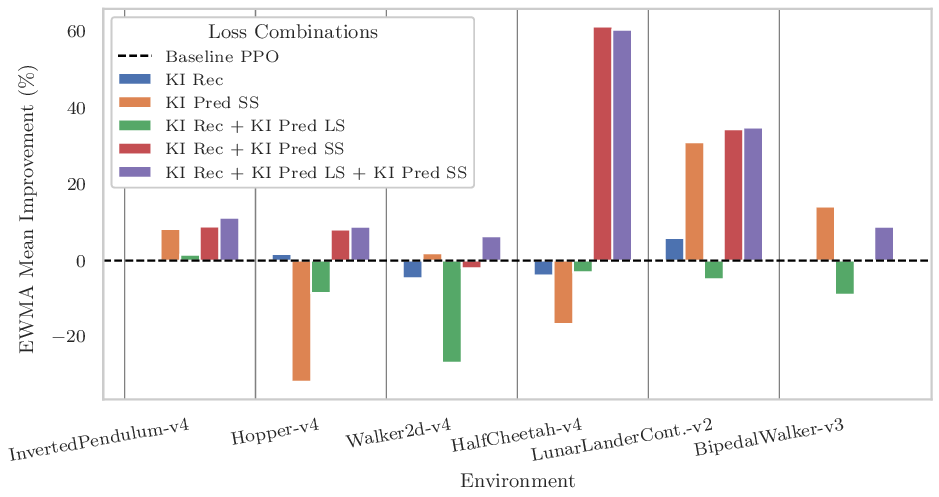}
  \caption{
    Percent improvement in mean final episode returns across different combinations of loss components compared to the baseline.
    Higher is better.
  }
  \label{fig:rel_impr_means_loss_combinations}
\end{figure}

\begin{figure}[tb]
  \centering
  \includegraphics[width=1.0\linewidth]{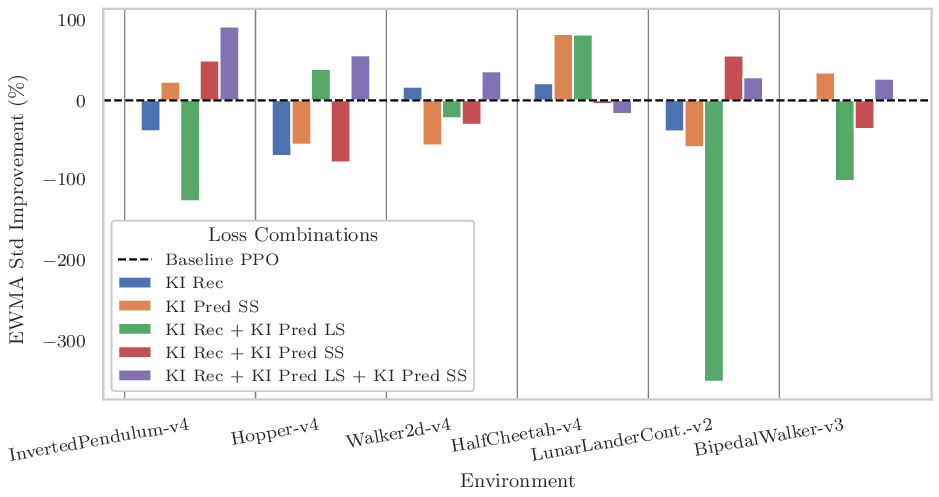}
  \caption{
    Percent improvement in standard deviation of final episodic returns across different combinations of loss components compared to the baseline.
    Higher is better.
  }
  \label{fig:rel_impr_stds_loss_combinations}
\end{figure}

These are the full results from the main paper (\Cref{sec:ablation}).
This section presents the numerical results and comparisons showing the impact of each loss function component.
The results are summarized in \Cref{tab:effect_losses} and also compared in percentage improvement in terms of both mean value and std.
of \gls{EWMA}, as shown in Figures \ref{fig:rel_impr_means_loss_combinations} and \ref{fig:rel_impr_stds_loss_combinations}.

\begin{table*}[htb]
  \tabEffectLosses{}
\end{table*}


\section{Extended Hyperparameter Analysis}\label{app:hypers}

To evaluate \gls{KIPPO}['s] sensitivity to various hyperparameters, we trained 300 model configurations, each with 4 random seeds across 6 environments, resulting in 7,200 trained models.
\Vref{tab:hypers_ranges} summarizes the explored hyperparameter values and ranges.
Unless otherwise noted, we report \emph{standardized} final episodic returns, measured relative to a \gls{PPO} baseline for each environment.

\begin{table}[htb]
  \tabHypersRanges{}
\end{table}

\subsection{Effect of Latent Dimension}\label{app:hypers:latent}

\begin{figure*}[htb]
  \centering
  \includegraphics[width=0.8\linewidth]{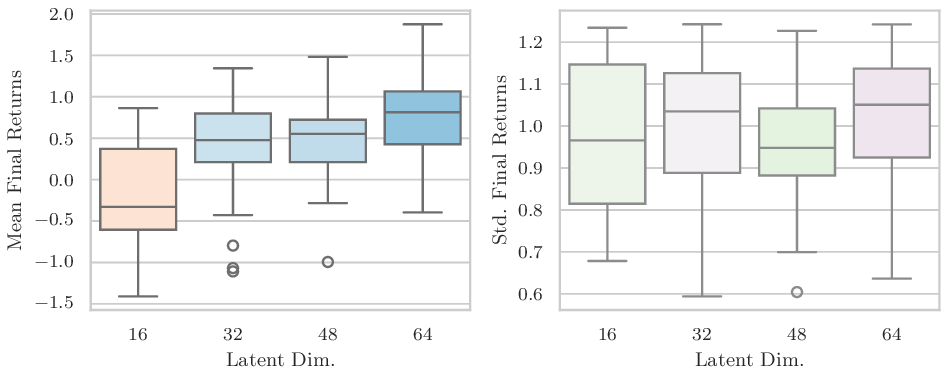}
  \caption{
    Effect of latent dimension on standardized final episodic returns.
    Larger dimensions generally yield higher returns and lower variability,
    although the optimal value depends on the task's complexity.
  }\label{fig:box_dyn_dim}
\end{figure*}

\Vref{tab:effect_latent_dim,fig:box_dyn_dim} examine the effects of varying the latent dimension among \{16, 32, 48\}.
Overall, dimensions of 32 or higher tend to boost returns and reduce variance, albeit at the cost of increased computational requirements.
For instance, \textit{HalfCheetah-v4} exhibits continued performance gains at larger dimensions, with noticeably lower variability at a dimension of 48.
In contrast, in \textit{InvertedPendulum-v4}, the most significant improvement occurs between 16 and 32, after which it returns to a plateau.
Consequently, the best-performing latent dimension can depend on the environment's complexity and dynamics.

\begin{table*}[htbp]
  \tabEffectLatentDim{}
\end{table*}

\subsection{Effect of Prediction Horizon}\label{app:hypers:horizon}

\begin{figure*}[htb]
  \centering
  \includegraphics[width=0.8\linewidth]{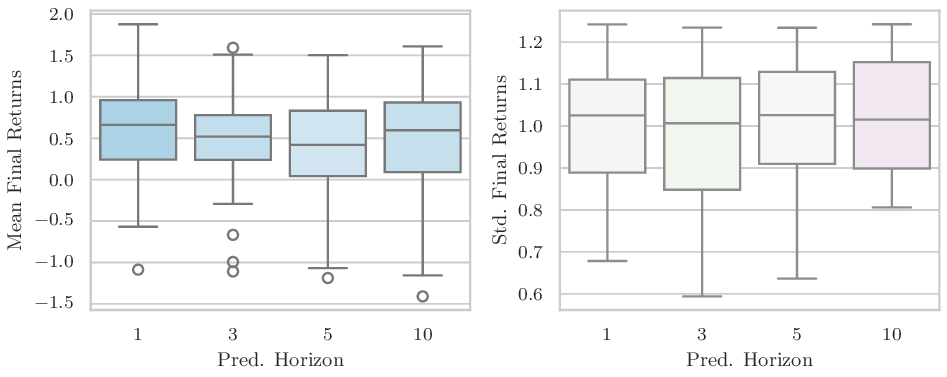}
  \caption{
    Effect of prediction horizon on standardized final episodic returns.
    Longer horizons can capture more distant dependencies but may amplify prediction errors.
  }\label{fig:box_horizon}
\end{figure*}

\Vref{tab:effect_horizon,fig:box_horizon} investigate horizons ranging from 1 to 10 steps.
Longer horizons capture extended dependencies but also risk compounding prediction errors.
In \textit{HalfCheetah-v4}, for example, a horizon of 10 achieves higher mean returns at the expense of slightly increased variance.
Conversely, \textit{Hopper-v4} exhibits diminishing returns and higher \gls{CTE} for horizons beyond 5.
In other cases, such as \textit{InvertedPendulum-v4} and \textit{LunarLanderContinuous-v2}, gains remain relatively consistent across all tested horizons, underscoring that optimal horizon length is often environment-specific.

\begin{table*}[htbp]
  \tabEffectHorizon{}
\end{table*}

\subsection{Effect of Loss Weights}\label{app:hypers:weights}

\begin{figure*}[htb]
  \centering
  \includegraphics[width=0.8\linewidth]{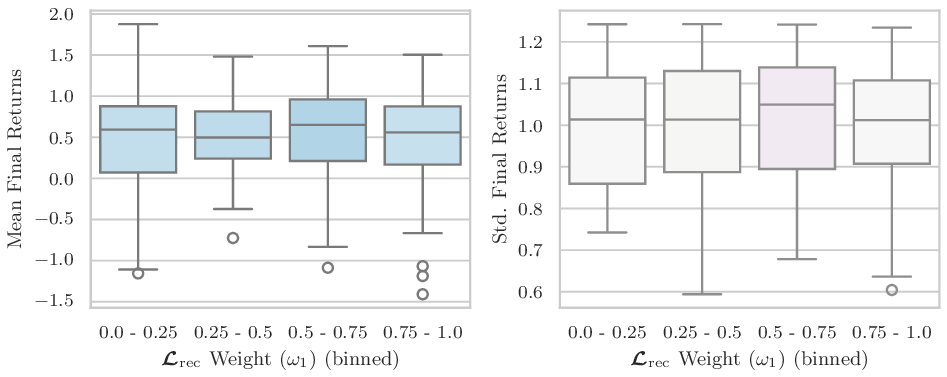}
  \caption{
    Effect of reconstruction loss weight on standardized returns.
  }\label{fig:box_loss_w_rec}
\end{figure*}

\begin{figure*}[htb]
  \centering
  \includegraphics[width=0.8\linewidth]{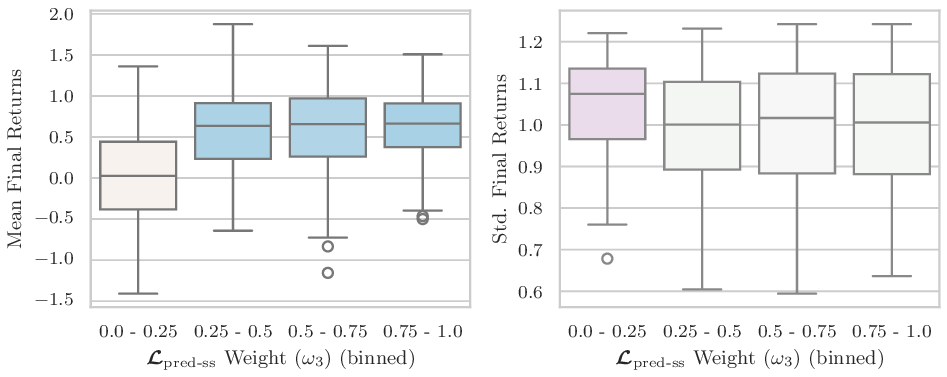}
  \caption{
    Effect of state-space prediction loss weight on standardized returns.
  }\label{fig:box_loss_w_pred}
\end{figure*}

\Vref{fig:box_loss_w_rec,fig:box_loss_w_pred} visualize how the three loss weights
(\gls{loss-weight-ki-rec}, \gls{loss-weight-ki-pred-ls}, and \gls{loss-weight-ki-pred-ss})
influence final episodic returns.
These weights were tested in increments of 0.05 from 0.00 to 1.00.

\begin{itemize}
  \item \textbf{Reconstruction weight (\gls{loss-weight-ki-rec}):}
        Most settings outperform the baseline, and values in the range of 0.25--0.50
        yield notably tighter interquartile ranges.
        Extremely high or low settings have minimal negative impact.
  \item \textbf{Latent-space prediction weight (\gls{loss-weight-ki-pred-ls}):}
        Values below 0.50 tend to yield higher mean returns and exhibit stable variance,
        though some environments benefit from a slightly higher weight.
  \item \textbf{State-space prediction weight (\gls{loss-weight-ki-pred-ss}):}
        Values above 0.25 generally outperform the baseline while maintaining similar variance.
        Configurations below 0.25 approach baseline performance but occasionally show slight increases in variability.
\end{itemize}

\begin{figure*}[htb]
  \centering
  \includegraphics[width=0.8\linewidth]{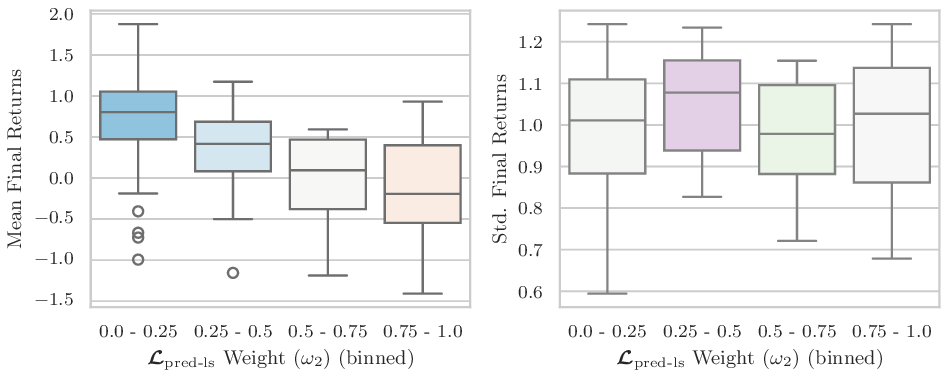}
  \caption{
    Effect of latent-space prediction loss weight on standardized returns.
  }\label{fig:box_loss_w_pred_enc}
\end{figure*}

\subsection{Effect of Miscellaneous Hyperparameters}\label{app:hypers:misc}

\begin{figure*}[htb]
  \centering
  \includegraphics[width=0.8\linewidth]{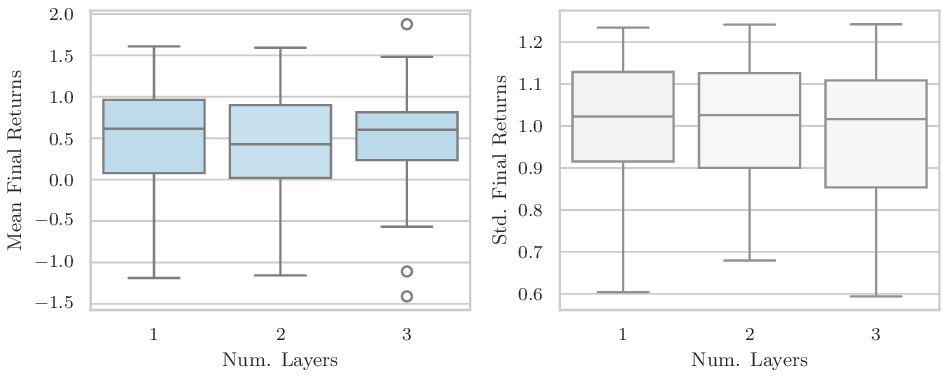}
  \caption{
    Effect of the autoencoder's network depth on standardized returns.
  }\label{fig:box_ae_n_layers}
\end{figure*}

\begin{figure*}[htb]
  \centering
  \includegraphics[width=0.8\linewidth]{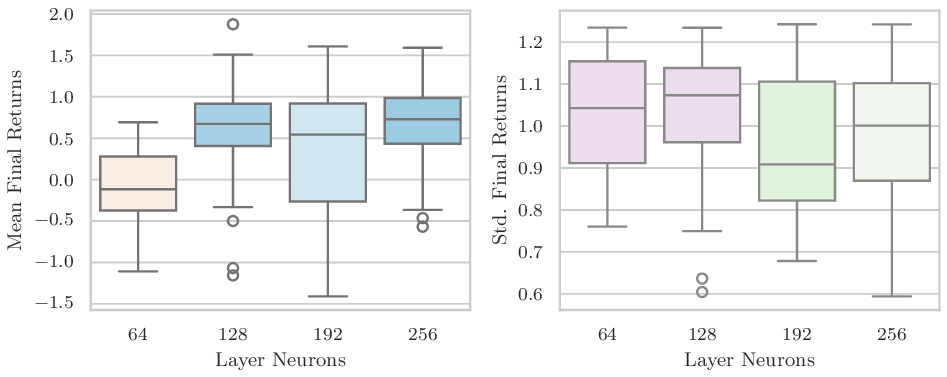}
  \caption{
    Effect of the autoencoder's feedforward dimension on standardized returns.
  }\label{fig:box_ae_dim_ff}
\end{figure*}

Finally, we explore architectural hyperparameters (\vref{fig:box_ae_n_layers,fig:box_ae_dim_ff}), including the number of network layers (1--3) and neurons per layer (64, 128, 192, 256).
Increasing the depth from 1 to 3 layers does not dramatically alter either mean returns or variance, indicating moderate robustness to network depth.
By contrast, network width has a more pronounced impact: fewer than 128 neurons per layer leads to underfitting in several environments, while 192--256 neurons improve performance and stabilize outcomes across the board.

Overall, these analyses highlight \gls{KIPPO}['s] broad robustness to different hyperparameter choices, while demonstrating that thoughtful tuning of critical parameters (e.g., latent dimension, horizon length, and network width) can yield substantial gains in performance and consistency across a variety of environments.


\section{Implementation Details}\label{app:impl}

\subsection{Rollouts and Data Collection}\label{app:impl:rollouts}

The rollout phase, detailed in~\Cref{alg:kippo_rollouts}, involves the agent interacting with the environment for a predefined number of steps, \(\gls{total-steps}\), to gather data for training the \gls{KIPPO} framework.
This phase efficiently collects data for both the \gls{PPO} algorithm and Koopman-inspired representation learning, enabling the joint optimization of the policy and latent-space representation.

At each timestep \(t\), the agent observes the current state, \(\gls{state}_t\), which is then encoded into its latent representation, \(\gls{state-obs}_t\), using the state encoder, \(\gls{state-encoder}\).
The policy, \(\gls{policy}_{\gls{params}}\), conditioned on \(\gls{state-obs}_t\), samples an action, \(\gls{action}_t\).
After executing this action in the environment, the agent receives a reward, \(\gls{reward}_t\), transitions to the next state, \(\gls{state}_{t+1}\), and observes a termination flag, \(\gls{done}_t\), indicating whether the episode has ended.
The algorithm stores this information, along with the action's log probability, \(\log\gls{policy}_t\), and the estimated state value, \(\gls{state-val}_t\), in corresponding buffers.

In addition to the standard \gls{RL} data, \gls{KIPPO} collects sequences of states, actions, and binary masks from the previous \(\gls{horizon}\) steps at each timestep \(t\).
These sequences, denoted as \(\gls{state}_{0:\gls{horizon}}\), \(\gls{action}_{0:\gls{horizon}-1}\), and \(\gls{binary-mask}_{1:\gls{horizon}}\), respectively, are used for computing the prediction losses.
The state sequence, of length \(\gls{horizon}+1\), uses the first element, \(\gls{state}_0\), as the initial state for prediction and the remaining \(\gls{horizon}\) elements as targets.
The action sequence, of length \(\gls{horizon}\), aligns with the state transitions.
The binary mask sequence, also of length \(\gls{horizon}\), is constructed based on episode termination flags, ensuring predictions are not made beyond episode boundaries.

Circular buffers (\(\gls{circ-buf-states}\), \(\gls{circ-buf-actions}\), \(\gls{circ-buf-dones}\)) efficiently manage the sliding window of recent experiences for sequence construction.
Storage tensors (\(\gls{stored-state-seqs}\), \(\gls{stored-action-seqs}\), \(\gls{stored-mask-seqs}\)) accumulate these sequences, preserving temporal structure for loss computation.

When an episode concludes, the environment is reset, and data collection continues until \(\gls{total-steps}\) steps have been collected.

\begin{algorithm*}[htb]
  \kippoRollouts{}
\end{algorithm*}

\subsection{Future State Prediction}\label{app:impl:pred}

\begin{algorithm*}[htb]
  \kippoPrediction{}
\end{algorithm*}

\subsection{Optimization Process}\label{app:impl:optim}

The optimization phase updates all networks and trainable parameters using the collected data.
\Vref{alg:kippo_optim} details this phase, highlighting the integration of \gls{KIPPO}['s] objectives with the \gls{PPO} algorithm.

The collected data is divided into mini-batches.
For each mini-batch, the algorithm computes the reconstruction loss, latent-space prediction loss, state-space prediction loss, and the \gls{PPO} loss, \(\gls{loss-ppo}\).
The actor and critic networks operate on the encoded states, enabling learning in the simplified latent space.
The \gls{nograd} operator ensures the state representations are optimized independently of the \gls{PPO} loss.

Future state prediction is performed efficiently using vector operations.
The weighted sum of the representation losses balances the different objectives.
The total loss combines these weighted losses with the \gls{PPO} loss.

Finally, gradients of the total loss are used to update the parameters via backpropagation using the Adam optimizer.
This process iteratively refines the latent-space representation, the policy, and the value function.

\begin{algorithm*}[htb]
  \kippoOptimization{}
\end{algorithm*}


\section{Latent Space Properties}\label{app:latent_space_props}

A non-linear system often requires a higher-dimensional representation to achieve linear dynamics.
Consider a system governed by the following dynamics:

\begin{equation}
  \begin{array}{lc}
    \frac{d}{dt}x_1 = x_1^2 \\
    \frac{d}{dt}x_2 = x_1 x_2 + x_2
  \end{array}
  \label{eq:example_non-linear_system}
\end{equation}

An appropriate coordinate transformation that enables linear evolution is:

\begin{equation}
  \begin{bmatrix}
    y_1 \\
    y_2 \\
    y_3
  \end{bmatrix}
  =
  \begin{bmatrix}
    x_1^2 \\
    x_2   \\
    x_1 x_2
  \end{bmatrix}
  .
  \label{eq:observable_funcs}
\end{equation}

Because some systems require a higher dimension to represent non-linear dynamics linearly, we encode to a space in a higher dimension than the state and action space.
Unlike \citet{song2021data}, we do not concatenate the original state with the encoded state, as this restricts the set of systems where linearization is possible.
Specifically, finding a linear representation of a non-linear system that includes the original state becomes impossible when the system has multiple fixed points or general attractors.
This limitation arises because linear systems (with a single fixed point at the origin) are not topologically conjugate to non-linear systems with multiple fixed points \citep{brunton2016koopman}.

In standard autoencoders, the encoding function is designed to be an embedding, which means it maps each point in the input space to a unique point in the latent space, ensuring a one-to-one correspondence.
This property is known as global injectivity.
By achieving global injectivity, the encoder creates a distinct and recoverable representation for each input state, allowing the decoder to reconstruct the original input accurately.
However, an encoder can also act as an immersion, characterized by \textit{local} injectivity.
In an immersion, each point in the input space has a neighborhood mapped injectively into the latent space.
This means that the mapping is one-to-one within a small region around each end.
However, an immersion may not maintain a one-to-one global mapping across the entire input space.
Lemma 1 from \citet{alford2022deep} extends this concept by stating that if an immersion is paired with a corresponding submersion that acts as its left inverse (in this case, in the form of a decoder), the composite function is the identity map on the input space.
This implies that the encoder is not just locally injective but globally injective over its entire domain, qualifying it as an embedding.
Under this condition, an encoder that is initially an immersion can indeed behave like a bijective embedding with respect to the subset of the latent space it maps onto, enabling a bi-directional, unambiguous mapping between the input space and its image in the latent space.

\end{document}